\documentclass[journal]{vgtc}

\ifpdf
  \pdfoutput=1\relax
  \usepackage{graphicx}
  \DeclareGraphicsExtensions{.pdf,.png,.jpg,.jpeg}
\else
  \usepackage{graphicx}
  \DeclareGraphicsExtensions{.eps}
\fi

\graphicspath{{figures/}{pictures/}{images/}{./}} 

\usepackage{microtype}
\PassOptionsToPackage{warn}{textcomp}
\usepackage{textcomp}
\usepackage{mathptmx}
\usepackage{times}

\usepackage{cite}
\usepackage{tabu}
\usepackage{booktabs}
\usepackage{amsmath}
\usepackage{amssymb}
\usepackage{multirow}
\usepackage{floatrow}
\usepackage{wrapfig}
\usepackage{picins}

\onlineid{0}

\vgtccategory{Research}
\vgtcpapertype{please specify}

\title{SAC-GAN: Structure-Aware Image Composition}

\author{Hang Zhou, Rui Ma, Ling-Xiao Zhang, Lin Gao, Ali Mahdavi-Amiri, and Hao Zhang}
\authorfooter{
\item
 Hang Zhou, Ali Mahdavi-Amiri, and Hao Zhang are with the School of Computing Science, Simon Fraser University, Canada. 
\item
 Rui Ma is with the School of Artificial Intelligence, Jilin University, China and Engineering Research Center of Knowledge-Driven Human-Machine Intelligence, MOE, China.
\item
 Ling-Xiao Zhang and Lin Gao are with the Institute of Computing Technology, Chinese Academy of Sciences, China. 
 \item
 Rui Ma is the corresponding author (Email: ruim@jlu.edu.cn) .
 \item
 This research is supported in part by a NSERC Discovery Grant (No.~611370) and National Natural Science Funds of China (62202199).
}

\shortauthortitle{Zhou \MakeLowercase{\textit{et al.}}: SAC-GAN: Structure-Aware Image Composition}

\abstract{
We introduce an end-to-end learning framework for {\em image-to-image composition\/}, aiming to plausibly compose an object represented as a cropped patch from an object image into a background scene image. As our approach emphasizes more on semantic and structural coherence of the composed images, rather than their pixel-level RGB accuracies, we tailor the input and output of our network with {\em structure-aware\/} features and design our network losses accordingly, with ground truth established in a {\em self-supervised\/} setting through the object cropping. Specifically, our network takes the semantic layout features from the input scene image, features encoded from the edges and silhouette in the input object patch, as well as a latent code as inputs, and generates a 2D spatial affine transform defining the translation and scaling of the object patch. The learned parameters are further fed into a differentiable spatial transformer network to transform the object patch into the target image, where our model is trained adversarially using an affine transform discriminator and a layout discriminator.
We evaluate our network, coined SAC-GAN, for various image composition scenarios in terms of quality, composability, and generalizability of the composite images. Comparisons are made to state-of-the-art alternatives, including Instance Insertion, ST-GAN, CompGAN and PlaceNet, confirming superiority of our method.
}


\CCScatlist{
 \CCScat{K.6.1}{Management of Computing and Information Systems}
{Project and People Management}{Life Cycle};
 \CCScat{K.7.m}{The Computing Profession}{Miscellaneous}{Ethics}
}

\teaser{
  \centering
  \includegraphics[width=0.95\linewidth]{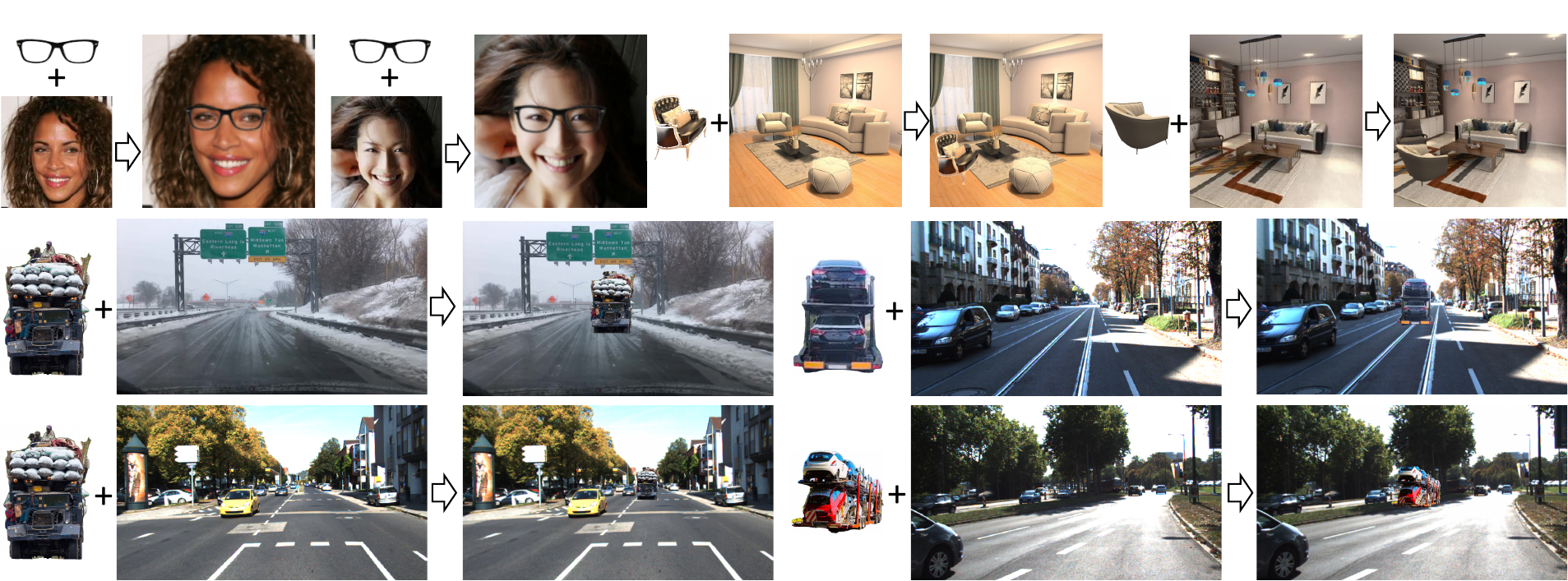}
  \caption{Our network SAC-GAN composes an object patch into a background image in a {\em structure-aware\/} manner by striving for semantic and structural (e.g., edge structures) coherence of the composed images. Composition results for faces and indoor/street scenes are shown, where for self-driving, rare vehicles can be added to images reflecting diverse weathers from KITTI~\cite{geiger2013vision} and BDD~\cite{yu2020bdd100k}, with the network trained on Cityscapes~\cite{cordts2016cityscapes} without rare vehicle augmentation, demonstrating the generalizability of our network.} 
  \label{fig:teaser}
}
\vgtcinsertpkg

\begin{document}

\maketitle
\vspace{-20pt}
\section{Introduction}
\label{sec:intro}

Compositional modeling is one of the most classical problems in computer graphics, covering such topics as image compositing~\cite{porter1984,brinkmann2008}, matting~\cite{wang2008_matting}, and structured assembly of 3D shapes~\cite{mitra_star13} and scene environments~\cite{egstar2020_struct}. Aside from emerging applications such as customized shopping, e.g., composing eyewear onto faces and furniture pieces into virtual homes (see top row of Fig.~\ref{fig:teaser}), a main area of applications for image composition is {\em data augmentation\/}~\cite{RaidaR,chen2021geosim,antoniou2018data,azadi2020compositional,DataAugSurvey,dwibedi2017cut,lin2018st}. Despite the recent explosion of visual content especially online images and video, the available data still does not fully address the need of modern machine learning. Take self-driving as an example, where most street scene data have been acquired by camera-mounted vehicles driven on real-world streets. This amounts to a random sampling over the immense spatio-temporal space of all possible scenarios a driver may encounter. Since the sample space is so large and diverse, current data collections still fall far short in terms of adequate coverage, e.g., of adverse weather conditions or rarely seen vehicles (see last two rows of Fig.~\ref{fig:teaser}) such as ambulances, tow trucks, car carriers, as well as seldom encountered traffic patterns involving accidents and constructions.

In this paper, we advocate the compositional approach to image manipulation and augmentation to serve the above application scenarios. Specifically, we aim to plausibly insert an object (e.g., a pair of glasses, a piece of furniture, a vehicle or pedestrian) represented as a {\em cropped patch\/} from an object image, into a target background scene, as shown in Fig.~\ref{fig:teaser}. In general, compositional image synthesis can lead to an exponential growth in the amount of possible image combinations. Furthermore, an object-oriented composition paradigm can allow more control over the semantics and structures of the resulting images than the composition based on pixel-level image synthesis \cite{lee2018context,hong2018learning}. Last but not least, compared to its 3D counterpart, e.g., compositional modeling of 3D structures~\cite{egstar2020_struct}, our {\em image-to-image\/} composition task has significantly more data to tap into to facilitate a learned solution, e.g., by learning from the abundant image datasets.

\begin{figure*}[!t]
\centering
\includegraphics[width=0.96\linewidth]{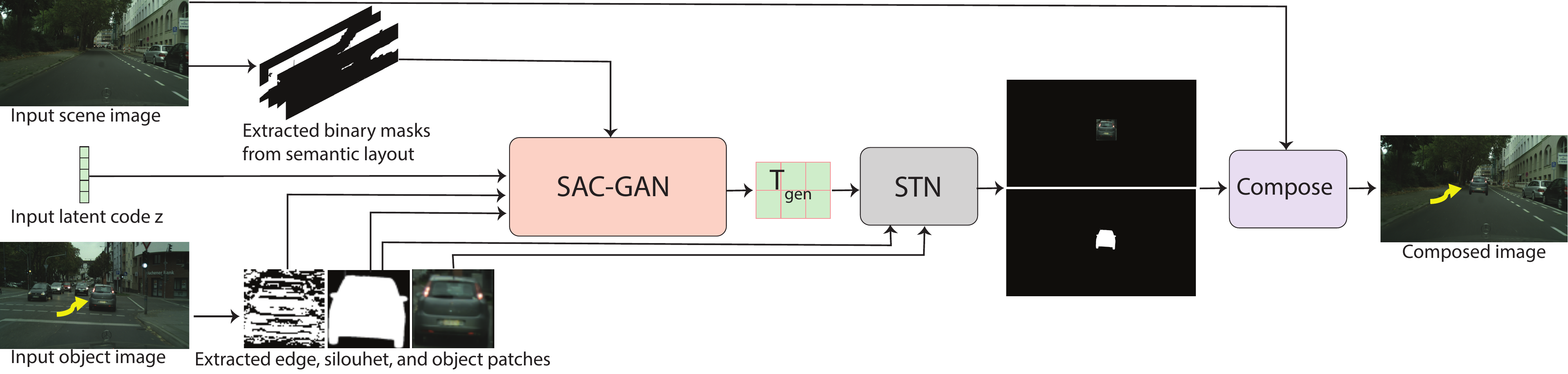}
\caption{\textbf{Network inference.} 
Given an object patch (yellow arrow) from an object image, a {\em new\/} background/target scene, and a latent code $z$, SAC-GAN predicts a 2D transform $T_{\mathrm{gen}}$, which is utilized by a spatial transformer network (STN), to produce a composed image. The composition is {\em structure-aware\/}, leveraging edge, silhouette, and semantic layout maps inferred from the inputs.}
\label{fig:inf_pipeline}
\end{figure*}

\begin{figure*}[!t]
\centering
 \includegraphics[width=0.96\linewidth]{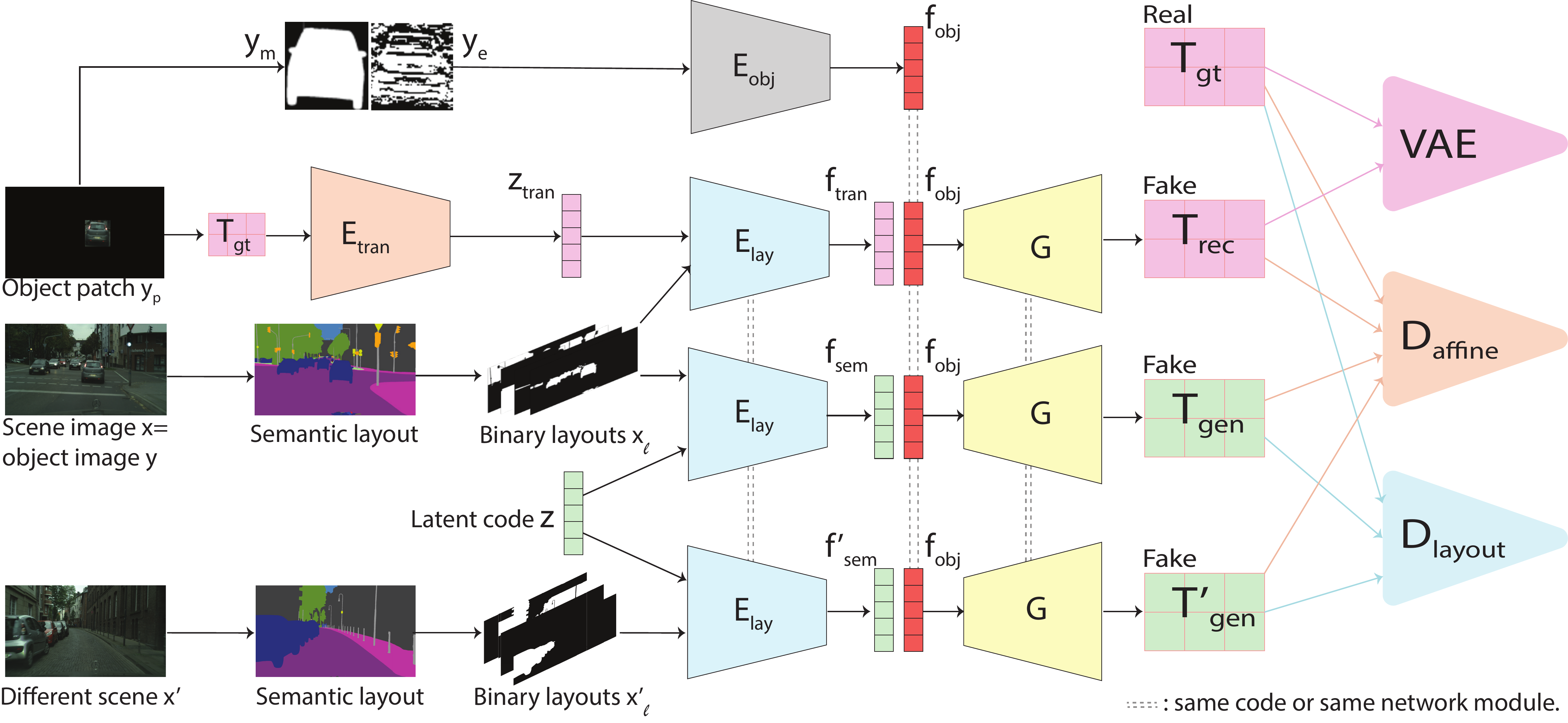}
\vspace{-4pt}
\caption{\textbf{Network architecture and training.}
During training, the object image ($y$) and the background/target image ($x$) are the {\em same\/}, allowing us to define GT transforms $T_{gt}$ with respect to cropped object patches ($y_p$). In addition, a different image ($x' \not= x$) is also processed to reinforce the transformation learning. Our object-aware VAE-GAN network feeds $T_{gt}$, a latent code $z$, and structure-aware features (semantic layouts $x_{l}$, object silhouette $y_m$, and edge map $y_e$) extracted from the input images ($x$, $y$, and $x'$) to their respective encoders for transformation ($\mathbf{E}_{trans}$), layout ($\mathbf{E}_{lay}$), and object patch ($\mathbf{E}_{obj}$). Finally, a shared decoder/generator $G$ takes the encoder outputs and produces three transformations, $T_{rec}$, $T_{gen}$, and $T'_{gen}$ for VAE and GAN learning via self-supervised loss and two discriminators $D_{affine}$ and $D_{layout}$.}
\label{fig:overflow}
\end{figure*}

The approach we introduce is an {\em end-to-end\/} neural network {\em adversarially\/} trained to compose an object patch into a target scene image. Our input transformation and network design have several distinguishing characteristics to reflect how we address the various challenges of our composition problem:
\begin{itemize}
\item First, our image composition is in essence an {\em object insertion\/}, and as such, we focus on learning how to {\em spatially transform\/} the input object patch to fit into a scene image; see Fig.~\ref{fig:inf_pipeline}. This is unlike conventional approaches to realistic neural image manipulation~\cite{dwibedi2017cut,lin2018st,cong2020dovenet} or transform~\cite{zhu2017unpaired,kim2017learning,yi2017dualgan,isola2017image}, including composition~\cite{azadi2020compositional,lin2018st}, which have focused on pixel-wise synthesis. Meanwhile, as we target to operate on the 2D domain so that the abundant image resources could be utilized, it raises significant challenges on obtaining plausible results with only 2D input as constraints. 
\item Second, we emphasize more on {\em semantic\/} and {\em structural coherence\/} of the composed image, rather than its pixel-level RGB realism \cite{niu2021} such as color harmonization and image blending. Since the scene semantics and structures provide more information about the relationships among the objects, they play a more important role than pixel-level coherence for learning a more plausible location and scale of the composed image. Hence, we extract {\em structure-aware\/} features such as silhouette, edge, and semantic layout maps, from the input images using classic edge detector and pre-trained segmentation models, and design the network architecture and losses to leverage these features.
\item Last but not least, our network is trained with {\em ground-truth\/} (GT) transformations, unlike previous works, e.g., ST-GAN~\cite{lin2018st}, whose generators are only judged by appearance-based realism discriminators. Yet, our GT does not require any paired data, either human-prepared or synthesized, e.g., via image inpainting~\cite{azadi2020compositional,zhang2020learning}. Instead, it is easily obtained under a {\em self-supervised\/} setting, through the object cropping process. Specifically, our network is trained to compose a cropped object back to the original image; see Fig.~\ref{fig:overflow}. 
\end{itemize}

Our network, coined SAC-GAN for {\em structure-aware composition\/}, predicts a 2D transformation to compose an object patch sampled from domain $Y$ and a target scene image sampled from domain $X$ to a composite image from domain $X$, as shown in Fig.~\ref{fig:inf_pipeline}. Our goal is to learn a mapping $F: (X, Y)\rightarrow X$ such that the joint distribution of the composite images from $F(X,Y)$ is indistinguishable from the marginal distribution $X$ using an adversarial loss. SAC-GAN is built on spatial transformer networks (STNs)~\cite{jaderberg2015spatial} and VAE-GAN~\cite{larsen2016autoencoding,lee2018context}. The choice of employing a VAE-GAN, rather a GAN or a VAE~\cite{verma2018generalized}, serves to avoid potential mode collapse or mode averaging issues with each respective model alone. Our model combines the strengths of VAEs and GANs by assembling them into a conditional generative model to learn 2D transforms from structure-oriented features extracted from the input images, rather than colors and textures which can exhibit variations and diversities to a greater extent. Though there are existing methods \cite{lee2018context,SiyuanZhou2022LearningOP} which also employ the VAE-GAN based architecture, they are either not object-aware (e.g., synthesis instead of using existing object masks \cite{lee2018context}) or need external composition supervision \cite{liu2021OPA} for learning the VAE \cite{SiyuanZhou2022LearningOP}.
Comparably, SAC-GAN is both object-aware and only needs self-supervision to learn the VAE-GAN. 

Fig.~\ref{fig:overflow} and the caption outline the architecture and training of our network, with more details in Section~\ref{sec:method}. For the VAE part, to train the network with GT transforms $T_{gt}$, we make the object and background/target images the same and learns $T_{rec}$ so that it transforms a default object patch (centered in image with size $64 \times 64$) to match its corresponding cropped patch. At the high level, our network takes the semantic layout features from the target scene image, features encoded from the edges and silhouette in the object patch, as well as a latent code as inputs, and generates a 2D spatial affine transform defining the translation and scaling (without rotation or shearing) of the object patch. The learned parameters are further fed into a differentiable STN to transform the object patch into the target image. For the GAN part, we train our model adversarially, using an affine transform discriminator and a layout discriminator to simultaneously ensure the learned 2D transforms and their induced composite scene layouts are realistic. Note $T_{gen}$ and $T'_{gen}$ are generated transforms to compose the object on the self-supervised scene/object image or a new scene, respectively. From another view, our network can be regarded as a conditional generative teacher-student network \cite{ZhengXu2018TrainingSA,TingYunChang2020TinyGANDB} that the student network (bottom branch in Fig. \ref{fig:overflow}) is trained to compose a given object patch onto a new scene by learning from the self-supervised teacher network.

We experiment with image composition tasks under different settings over well-known datasets, such as vehicle insertion on Cityscapes~\cite{cordts2016cityscapes}, glasses insertion on CelebAHQ~\cite{liu2015faceattributes} and furniture insertion on 3D-FUTURE~\cite{fu20213d}. Qualitative and quantitative comparisons were made to Instance Insertion~\cite{lee2018context}, ST-GAN~\cite{lin2018st}, CompGAN~\cite{azadi2020compositional}, and PlaceNet \cite{zhang2020learning}, as representative methods for learning-based object insertion, and the results demonstrate superiority of SAC-GAN. We also show the generalizability of SAC-GAN by composing Google searched cars into new test images in KITTI~\cite{geiger2013vision}, BDD~\cite{yu2020bdd100k}, and Mapillary Vistas~\cite{neuhold2017mapillary}.
Moreover, we show the potential of SAC-GAN for data-augmentation via quantitative experiments on semantic segmentation. At last, we conduct ablation studies to validate our network designs, demonstrating that simpler versions, e.g., only using a VAE/GAN loss, do not attain the same composition quality.  

In summary, our contributions are as follows:
\begin{itemize}
\item We propose a novel end-to-end learning framework for structure-aware image-to-image composition.
By focusing on the semantics and structures of the scene and object, we can directly operate on the 2D domain and produce composition with more plausible object locations and scales than existing methods which focus on pixel-level coherence. 
\item To the best of our knowledge, we are the first to employ a self-supervised and object-aware VAE-GAN architecture to learn transformations for the given object mask and the target scene. The pipeline is generalizable and scalable for learning image composition on other datasets as long as pre-trained models for extracting scene layouts and object masks are available. 
\item We perform extensive qualitative and quantitative experiments and show SAC-GAN outperforms SOTA image composition baselines in terms of quality, generalizability and the potential to be used for data augmentation.
\end{itemize}

\section{Related Work}
\label{sec:related}

We cover related works on image composition and generative image synthesis with more focus on object insertion. 

\textbf{Location-specified object insertion.}
Composing 3D objects into 2D images has been studied in \cite{debevec1998rendering,karsch2011rendering,karsch2014automatic,kholgade20143d,Alhaija2018IJCV}.
To insert a virtual 3D object into the image, the geometry and lighting of the scene are estimated~\cite{karsch2014automatic,yi2018faces,einabadi2021deep}, and the composite image can be generated by rendering the object at a specified location following the graphics pipeline. Compared to 3D objects, there are far more resources of available 2D object patches from existing images, which can enable the object insertion in the image-to-image composition manner. Early 2D-based methods such as alpha matting~\cite{smith1996blue} and Poisson image editing~\cite{perez2003poisson} stitch a foreground object into a background image by blending a local transition region. By retrieving object patches from a vast image-based object library, Lalonde et al.~\cite{lalonde07} design an interactive system allowing users to insert object patches with similar camera pose and lighting into the background scene. Although the above 3D and 2D methods can produce visually appealing composite images, the objects' location all need to be specified beforehand.

\textbf{Learning-based image composition.}
As summarized in Niu et al. \cite{niu2021}, most of existing image composition methods focus on the RGB pixel-level realism for object placement, color harmonization, shadow generation and image blending. Focusing on human instance composition, Tan et al.~\cite{tan2018and} learn to predict the potential location and scaling for insertion based on the input image and its semantic layout, and perform context-based image retrieval to find the most suitable person segments for composition. To automatically determine the location and other transformation such as rotation and scaling of more general object categories, Lin et al.~\cite{lin2018st} propose ST-GAN to learn the 2D geometric warp parameters of the foreground object patch based on STN \cite{jaderberg2015spatial}, which is a differentiable module to predict the transformation of a given image. CompGAN \cite{azadi2020compositional} further considers the interaction (e.g., occlusion) between objects and proposes a self-consistent composition-by-decomposition network to compose a pair of objects. Zhang et al.~\cite{zhang2020learning} propose PlaceNet based on directly learning the locations of the inserted object with the paired target and inpainted scene images. Besides geometry, appearance (e.g., color and brightness of the object patches) is adjusted in \cite{zhan2019spatial,chen2019toward} and the object shadows are predicted in \cite{wang2020people,wang2021repopulating} to improve the realism of the composition.

To evaluate the pixel-level realism of the composite images, the Object Placement Assessment (OPA) dataset \cite{liu2021OPA} in which the composite images with GT true/false labels are created and simple binary classifiers can be trained to predict whether an image is composite or not. Based on the composite images (with known GT object transform) in OPA, Zhou et al. \cite{SiyuanZhou2022LearningOP} propose GracoNet, a VAE-GAN based dual-path framework to solve the object placement problem via graph completion. Since their graph does not correspond to the semantic concepts in the image, i.e., the nodes are from uniformly partitioned patches and the edges are based on neighboring relationships, GracoNet still mainly focuses on the pixel-level coherence of the composite image. In contrast, we emphasize more on the structure coherence between the composed object and the scene when learning the transformation by encoding edge+silhouette information in the object patch and incorporating semantic scene layouts into our network. Hence, a natural integration in terms of object/scene semantics and structures during the composition can be ensured. Moreover, instead of being trained with external supervision like GracoNet\cite{SiyuanZhou2022LearningOP}, our network is trained in a self-supervised manner by obtaining the GT transforms based on the object cropping and its corresponding non-cropped scene image. This provides a stronger supervision and allows more stable training than GAN-based methods such as ST-GAN~\cite{lin2018st}, CompGAN~\cite{azadi2020compositional} and PlaceNet~\cite{zhang2020learning} which perform adversarial training by only optimizing the realism of the composite images.

\textbf{Semantic composition and synthesis.}
Instead of operating in the image domain, object insertion can be performed in the semantic domain \cite{lee2018context,hong2018learning}, e.g., on semantic label maps obtained from the pre-trained semantic segmentation models such as \cite{yu2018bisenet, tao2020hierarchical, ding2020semantic, gu2022multi}. 
The composite image can then be generated by GAN-based image-to-image translation \cite{isola2017image,zhu2017unpaired,yi2017dualgan,wang2018high,park2019semantic}. For example, given an input semantic map, Lee et al. \cite{lee2018context} introduce context-aware instance insertion which learns to predict plausible locations and synthesize the shapes of object instance masks by considering the scene context and generate multiple new semantic maps with objects inserted. Beyond insertion by synthesizing object instance masks, the semantic maps of the whole scene can also be generated following conditions such as object class proportions \cite{le2021semantic}. From the semantic maps, Wang et al. \cite{wang2018high} employ object instance information and enable object-level semantic manipulation and synthesis of high-resolution images. Ntavelis et al. \cite{ntavelis2020sesame} propose a user-guided semantic image editing method for adding, manipulating or erasing objects. Yang et al. \cite{yang2022modeling} design a layer-to-image generative Transformer-based model with focal attention to separately model image compositions on object-level and patch-level by distinguishing between object and patch tokens. These methods \cite{lee2018context,le2021semantic,wang2018high} mainly target instance-aware masks and image synthesis rather than image-domain composition. In the semantic domain, comparing to \cite{lee2018context}, we predict the transformation of a given object patch instead of synthesizing a new object mask. In this way, our network can produce more object-scene compatible transformation by incorporating the object's own structure information in the network learning.

\textbf{Object-aware neural scene composition.}
With the recent emergence of neural rendering \cite{tewari2020state}, object insertion and scene composition have been tackled by leveraging neural renderers for object appearance \cite{bhattad2020cut} and novel views \cite{chen2021geosim}, or learning object-aware neural scene representation \cite{Niemeyer2020GIRAFFE,ost2021neural,guo2020osf,yang2021learning}. One representative work is GeoSim \cite{chen2021geosim}, which first extracts objects as reconstructed 3D assets from reference videos, and then synthesizes new videos by performing novel-view neural rendering of the objects at locations sampled from the given 3D layout (i.e., HD maps). While such neural scene composition works as GeoSim can produce promising appearance and geometry in the composite images or videos, they usually require multi-view images of the object to train the networks. In contrast, we only use single-view object observation to train our model and focus on placing object patches with proper transformation rather than creating new object appearances or views at specified location.

\section{Method}
\label{sec:method}
Given an object image $y$ and a scene image $x$, our goal is to learn a transformation matrix $T_{gen}$ that successfully places an object patch $y_p$ existing in $y$ into the scene image $x$ (see Fig. \ref{fig:teaser}). Transformation $T_{gen}$ is applied on $y_p$ and its corresponding object mask $y_m$ via a spatial transformer (STN)~\cite{jaderberg2015spatial}. The outcomes are then used for composition which is a cut-and-paste operation:
\begin{equation}
\begin{aligned}
\label{eqn:cut-paste}
h(x, y, T_{gen})=&x\oplus T_{gen}(y)
          =T_{gen}(y_m\circ y_p)+(1-T_{gen}(y_m))\circ x,
\end{aligned}
\end{equation}
where $\oplus$ denotes mask blending and $\circ$ is the Hadamard product.

In our work, we only consider isotropic scaling and 2D translation of the object patches for $T_{gen}$.
We ignore the in-plane rotation by assuming the given object patch is aligned with the scene in the upright direction. Moreover, we do not model the out-of-plane rotation and the perspective transformation, though they can enable more realistic and diverse compositions. The reason is we only rely on object observation from a single view rather than the multiple views as in GeoSim \cite{chen2021geosim}, and only translation and scaling can be learned from the self-supervised object cropping process.
Specifically, the transformation matrix $T_{gen}$ is formulated as
\begin{equation}
T_{gen}=\left[\begin{matrix}
   s & 0 & t_x \\
   0 & s & t_y
  \end{matrix}\right],
\end{equation}
where $s$ is the scale and $t_x$, $t_y$ define the 2D translation. 

\subsection{Network Architecture and Loss Functions}
\label{sec:network}

At the high level, SAC-GAN follows the general VAE-GAN \cite{larsen2016autoencoding} architecture and
has three encoders: the spatial-transformer based transformation encoder $\mathbf{E}_{tran}$, the CNN based scene layout encoder $\mathbf{E}_{lay}$ and object structure encoder $\mathbf{E}_{obj}$ which encode the semantics and structures (in the form of different masks) for the scene and object. As in VAE-GAN, our decoder and generator is a shared network that reconstructs/generates the object's transformation. Generally, SAC-GAN is a conditional model that employs both the generative ability of VAEs and GANs through self-supervised and adversarial learning. Technically, only employing the GAN for transformation generation may result in mode collapse, ignoring conditions on the input patch and the scene image. As VAEs are known to enable more stable feature generations than GANs, adding a VAE loss can help alleviate the mode collapse.

\textbf{Transformation encoder.} 
The input to $\mathbf{E}_{tran}$ is the GT transformation $T_{gt}$ for the $x-y$ pair. Note that in training, we use the same image for the scene image $x$ and object image $y$. Hence, $T_{gt}$ is directly available by computing the transformation between the object patch and the scene image. The output of $\mathbf{E}_{tran}$ is a latent code $z_{tran}$ which encodes the GT transformation.

\textbf{Transformation-aware layout encoder.}
Next, we pass $z_{tran}$ along with a set of binary layout masks $x_l$ to $\mathbf{E}_{lay}$ and obtain a \textit{transformation-aware} semantic scene feature $\mathbf{f}_{tran}$. By integrating $z_{tran}$ into the scene feature, we encourage the decoder to automatically recover the object transformation given $\mathbf{f}_{tran}$ and the object feature $\mathbf{f}_{obj}$ (see details in the decoder). Separately, to enable the generation of transformation in the inference stage, $\mathbf{E}_{lay}$ also receives a latent code $z$ sampled from the standard Gaussian distribution and produces a generated semantic scene feature $\mathbf{f}_{sem}$ using $z$ and $x_l$.
For the scene layout masks $x_l$, we can either use ground truth masks provided by the training dataset or the segmentation masks predicted by off-the-shelf pre-trained segmentation networks (see Sec. \ref{sec:impl_details}) and turn the semantic masks of each class (e.g., road) into a set of binary masks. With semantic masks, the network focuses more on the semantic relationship between the scene image and the object patch and the use of separate binary masks can facilitate learning structure-aware features in a scene.

\textbf{Object structure encoder.}
Given an object image $y$, since the semantic or instance masks may also be available from our training dataset, we just select one object with a distinguishable and intact mask $y_m$ and use its bounding box to get a cropped object patch $y_p$. During testing, when the object masks are not available (e.g., for Google images), we apply off-the-shelf semantic segmentation algorithm as done for the scene images to extract the masks and select one intact target object as the input $y_p$. In addition, an edge map $y_e$ is extracted from $y_p$ via the classic Sobel operator. $\mathbf{E}_{obj}$ takes $y_m$ and $y_e$ as inputs, and outputs an object feature $\mathbf{f}_{obj}$. Comparing to only using silhouettes in mask $y_m$, edge map $y_e$ contains more semantic and structure information contained in the interior of the object patch, which can facilitate more realistic and orientation-sensitive compositions for objects like vehicles and furniture. 

\textbf{Shared transformation decoder-generator.}
In our VAE-GAN based architecture, the VAE decoder and the GAN generator are collapsed into one network $\mathbf{G}$ by sharing parameters and joint training. For the 2D transform reconstruction of the VAE branch, $\mathbf{G}$ takes the scene and object feature $(\mathbf{f}_{tran}$,$\mathbf{f}_{obj})$ as input and tries to reconstruct $T_{gt}$ by producing $T_{rec}$. For the GAN branch, the generator $\mathbf{G}$ takes the pair of generated scene feature and the object feature $(\mathbf{f}_{sem}$,$\mathbf{f}_{obj})$ and generates the 2D transformation $T_{gen}$. Note that the scene feature $\mathbf{f}_{tran}$ and $\mathbf{f}_{sem}$ already embed the transformation information and the decoder/generator will try to recover/generate the transformation for the object represented by $\mathbf{f}_{obj}$. The direct output of $\mathbf{G}$ is a 3-dimensional vector containing the scaling value $s$ and translation $t_x, t_y$, from which the transformation matrix can be built.

\textbf{Discriminators.} 
To ensure that the generated 2D transformations are diverse and reasonable with respect to locations and scales, we employ an adversarial affine transform discriminator $\mathbf{D}_{\it{affine}}$ which aims to distinguish whether the transformation parameters are realistic or not. Specifically, $\mathbf{D}_{\it{affine}}$ discriminates three 2D transformation matrices from the GT $T_{gt}$, including the reconstructed $T_{rec}$, the generated $T_{gen}$ and another generated $T_{gen}'$ which aims to realistically compose the object $y_p$ into to a new scene image $x'$.
To compute $T_{gen}'$, we pass the same latent vector $z$, the new scene layout $x'_l$ and $y_m$, $y_e$ to the encoders $\mathbf{E}_{tran}$, $\mathbf{E}_{lay}$, $\mathbf{E}_{obj}$, respectively. Moreover, to encourage the generator to focus on the relationship between semantic layouts and object patches, we design an adversarial layout discriminator $\mathbf{D}_{layout}$ so that the preferred locations are determined according to the global semantic context of the input. $\mathbf{D}_{layout}$ discriminates the generated $T_{gen}$ and $T_{gen}'$ from the GT $T_{gt}$ by discerning the original and composed scene layout maps.

\textbf{Loss functions.} 
Our training loss is formulated as 
\begin{equation}
\mathcal{L}_{VAE-GAN}=\mathcal{L}_{GAN}+\alpha\mathcal{L}_{VAE},
\end{equation}
where $\alpha$ is a weighting factor. We employ a VAE to regularize the affine transform encoder $\mathbf{E}_{tran}$ by imposing a Gaussian prior over the latent distribution $p(z)$. Concretely, our VAE loss $\mathcal{L}_{VAE}$ is defined by 
\begin{equation}
\label{eqn:04}
\mathcal{L}_{VAE}=\mathbb{E}_{z_{tran}}\|T_{rec}-T_{gt}\|_2^2+\mathrm{KL}(z_{tran}||z),
\end{equation}
where $\mathrm{KL}(\cdot)$ is the Kullback-Leibler divergence, $z_{tran}$ is an encoded vector from parameters of $T_{gt}$, and $\mathbb{E}_{(\cdot)}\triangleq\mathbb{E}_{(\cdot)\sim p_{data}(\cdot)}$ represents the expectation when the input variable follows distribution $p_{data}(\cdot)$.

Our GAN loss $\mathcal{L}_{GAN}$ consists of an adversarial affine transform loss $\mathcal{L}_{GAN}^{\it{affine}}$ and a layout loss $\mathcal{L}_{GAN}^{layout}$: 
\begin{equation}
\label{eqn:05}
\mathcal{L}_{GAN}=\mathcal{L}_{GAN}^{\it{affine}}+\beta\mathcal{L}_{GAN}^{layout}, 
\end{equation}
\begin{equation}
\label{eqn:06}
\begin{aligned}
\mathcal{L}_{GAN}^{\it{affine}}=&\mathbb{E}_{z_{tran}}\log \mathbf{D}_{\it{affine}}(T_{gt}) +\mathbb{E}_{z_{tran}}\log (1-\mathbf{D}_{\it{affine}}(T_{rec}))\\
&+\mathbb{E}_{z}\log (1-\mathbf{D}_{\it{affine}}(T_{gen})) +\mathbb{E}_{z}\log (1-\mathbf{D}_{\it{affine}}(T_{gen}')),
\end{aligned}
\end{equation}
\begin{equation}
\label{eqn:07}
\begin{aligned}
\mathcal{L}_{GAN}^{layout}=&\mathbb{E}_{x,y}\log \mathbf{D}_{layout}(x_l\oplus T_{gt}(y_m))\\
&+\mathbb{E}_{x,y,z}\log (1-\mathbf{D}_{layout}(x_l\oplus T_{gen}(y_m)))\\
&+\mathbb{E}_{x,y,z}\log (1-\mathbf{D}_{layout}(x'_l\oplus T_{gen}'(y_m))), 
\end{aligned}
\end{equation}
where $\beta$ is a weighting factor and $\oplus$ is the same operation defined in Equation \ref{eqn:cut-paste} while it works on the scene and object masks here. In general, $\mathcal{L}_{GAN}^{\it{affine}}$ focuses on predicting a realistic 2D transformation and $\mathcal{L}_{GAN}^{layout}$ enables the semantic and structure constraints from the scene and object. The total loss encourages the shared transformation decoder-generator $\mathbf{G}$ to learn the relationship between the scene and object in an object-aware and self-supervised manner.

\begin{figure*}[t]
\centering
\includegraphics[width=0.99\linewidth]{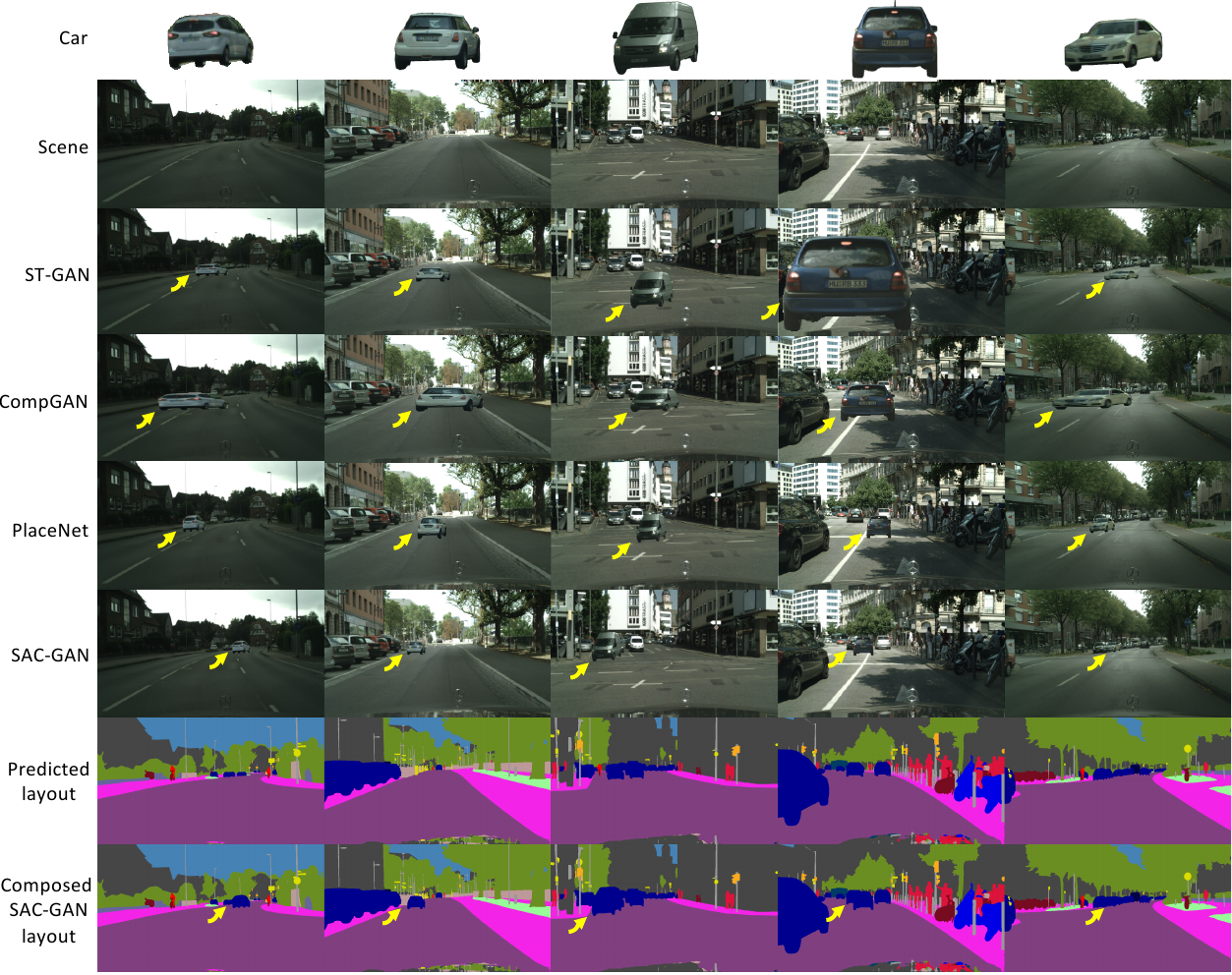}
\caption{Qualitative comparison of car insertion with ST-GAN~\cite{lin2018st}, 
CompGAN~\cite{azadi2020compositional} and PlaceNet~\cite{zhang2020learning} on Cityscapes. For a clear comparison, the inserted objects are marked with yellow arrows. Comparing with ST-GAN and CompGAN, SAC-GAN generates better object scales. Comparing with PlaceNet, SAC-GAN generates better semantic-aware locations, e.g. the locations of the left 3 samples of SAC-GAN are more lane-aware than PlaceNet, and the scales of the right 2 samples of PlaceNet appear smaller w.r.t their inserted locations. In addition, the SAC-GAN layouts show similar semantic distribution with GT layouts, being attributed to the constraints from the adversarial affine and layout losses. For all figures in the paper, please zoom in to see more visual details. }
\label{fig:compare}
\end{figure*}

\subsection{Training and Implementation Details}
\label{sec:impl_details}
In the following, we provide the details of our training scheme, datasets, network architecture and implementation details.

\textbf{Self-supervised training scheme.}
In training, scene image $x$ and object image $y$ are the same. Thus, our goal of learning object composition is equivalent to the self-composition of scene objects. By using the differentiable STN for object transformation, we can train our network in an end-to-end manner. For annotating object instance masks, we select a non-occluded object from the instance IDs of segmented images or pick an intact object from the annotated labels. The object patches are cropped and scaled to $64\times 64$ with their original aspect ratios. The scene image and segmented object naturally form a training pair which can be used to train the network in a self-supervised manner. The same scene image may be used to train different models when the object in a different category is selected.

\textbf{Training datasets.}
Separate models have been trained on different datasets for inserting different object categories. We use images from Cityscapes \cite{cordts2016cityscapes} training set to learn the car insertion model and use the Cityscapes-Aachen subset which contains more city views for pedestrians and street lights/signs. To further verify the generalizability and show the application of SAC-GAN, we have also trained models for inserting glasses on
CelebAHQ \cite{liu2015faceattributes} and chairs on 3D-FUTURE \cite{fu20213d}. As our network relies on the semantic layout masks, we apply the existing pre-trained Hierarchical Multi-Scale Attention (HMSA) network \cite{tao2020hierarchical} for semantic segmentation of street scenes and BiSeNet \cite{yu2018bisenet} for segmenting faces. For indoor scenes from 3D-FUTURE, we directly use the provided GT masks which are produced by the rendering process. In the experiments (Sec. \ref{sec:eval} and Sec. \ref{sec:app}), we show our method can adapt to both predicted semantic layout and the GT layout masks.

\textbf{Details of network architecture.} 
For the two encoders $\mathbf{E}_{lay}$ and $\mathbf{E}_{obj}$, we utilize CNNs with instance normalization to extract scene layout and object features. For the transformation encoder $\mathbf{E}_{tran}$, we employ the convolutional layers with 1 $\times$ 1 kernel to learn the mean $\mu$ and logarithmic variance $\ln{(\sigma^2)}$ of the latent code $z_{tran}$, and then obtain the final $z_{tran}$ by:
\begin{equation}
\label{eqn:vae}
z_{tran}=\mu+\sigma\cdot\epsilon, 
\end{equation}
where $\epsilon$ is sampled from standard Gaussian distribution.
For decoder-generator $\mathbf{G}$, we utilize fully connected layers to predict the 2D transformation parameters. 
For $\mathbf{D}_{affine}$, 1 $\times$ 1 convolutional layers are also used to map the 6-dimensional input to $32$ channels and then shrink it to one dimension.
For $\mathbf{D}_{layout}$, a PatchGAN-style~\cite{zhu2017unpaired} architecture with convolutional layers is used. 

{\textbf{Implementation details.}}
We have implemented our network in PyTorch and used Adam optimizer with StepLR scheduler (step size: $30$, gamma: $0.75$), an initial learning rate of $0.0002$, and a batch size of $1$ for training. For the losses, we set $\alpha$ to $1.5$ and $\beta$ to $1$. 
In all experiments, we generate layouts and images with resolution $960 \times 540$. We train separate models for cars, street lights, glasses and chairs. 
Each model is trained for $100$ epochs using one NVIDIA TITAN RTX GPU. 
Training takes about $12$ hours for cars, $4$ hours for pedestrians or street lights, $7$ hours for glasses and $10$ hours for chairs.
For inference, excluding the time for segmentation of the object and scene images, our network takes about $1.91$ seconds to generate a transformation.

\section{Results and Evaluation}
\label{sec:experiments}

In this section, we provide qualitative and quantitative results and experiments to demonstrate the effectiveness of SAC-GAN. We examine its performance in various object insertions such as cars or pedestrians in road scenes, or glasses on facial images, and furniture pieces in indoor scenes. We compare our results with state-of-the-art object insertion methods. We also perform ablation studies to assess the efficacy of our design. Our code and data is avaliable at https://github.com/RyanHangZhou/SAC-GAN.

\subsection{Evaluation and Comparison}
\label{sec:eval}

We compare SAC-GAN with four prior methods, ST-GAN \cite{lin2018st}, CompGAN \cite{azadi2020compositional}, PlaceNet \cite{zhang2020learning} and Instance Insertion \cite{lee2018context}, on car or pedestrian insertion. The first three are the representatives for learning-based direct composition methods and the last one is specific for semantic composition. Therefore, we compare with existing methods both on the pixel-level (Sec. \ref{sec:percep_eval}) and semantic-level object placement (Sec. \ref{sec:semantic_eval}).

For ST-GAN, we use their model trained on Cityscapes at $144 \times 144$ resolution and CompGAN, their model is trained at $256 \times 128$ resolution. For PlaceNet, we train their model at $960 \times 540$ resolution. For a fair comparison, we \textit{randomly} pick object-scene pairs from the Cityscapes validation set for composition, and recompose them with the learned object placement at $960\times 540$ resolution. For Instance Insertion \cite{lee2018context}, since it is only designed for semantic composition rather than direct object composition, to allow side-by-side perceptual comparison, we first generate object-inserted layouts by their network trained on Cityscapes at $1024 \times 512$ resolution. Note in the original implementation of Instance Insertion, it does not perform object composition directly. Instead, it first predicts the location and size of a bounding box for insertion, and then generates a mask of the object (e.g., a car) at the predicted location. To conduct a fair comparison, we apply Instance Insertion to predict the bounding box for insertion and then fit the input car or pedestrian mask into the box. Afterwards, we feed the layouts to Pix2PixHD \cite{wang2018high} for composite image generation.

\floatsetup[table]{capposition=top}
\begin{table}
\setlength{\tabcolsep}{1pt}
\begin{center}
\caption{Perceptual quality evaluation (left) and comparison via object detection (right) results for inserting cars or pedestrians on Cityscapes. Human score (HS) is the \% of participants who prefer SAC-GAN results over the baseline (higher number means lower preference of the compared method), Fr\'{e}chet inception distance (FID) metric assesses the quality of images created by a generative model, and precision at certain IoU threshold measures the plausibility of the composed objects via the object detection test.}
\label{tab:perceptual_iou}
\scalebox{0.92}{
\begin{tabular}{lccc|ccc}
\hline
Method & Type & HS ($\%$)$\uparrow$ & FID$\downarrow$ & $\mathrm{P_{IoU75}}\uparrow$ & $\mathrm{P_{IoU80}}\uparrow$ & $\mathrm{P_{IoU85}}\uparrow$\\
\hline
\multirow{2}*{Instance Insertion~\cite{lee2018context}} & Car & $93.6$  & $75.4$ & $0.72$ & $0.66$ & $0.58$\\
& Pedestrian & $95.1$ & $68.8$ & $0.35$ & $0.28$ & $\mathbf{0.18}$ \\
\hline
ST-GAN~\cite{lin2018st} & Car & $73.7$ & $74.3$ & $0.49$ & $0.46$ & $0.37$\\
\hline
CompGAN~\cite{azadi2020compositional} & Car & $84.6$ & $40.3$ & $0.89$ & $0.87$ & $0.76$\\
\hline
PlaceNet~\cite{zhang2020learning} & Car & $82.0$ & $36.4$ & $0.62$ & $0.44$ & $0.25$ \\
\hline
\multirow{2}*{SAC-GAN} & Car & --- &  $\mathbf{24.3}$ & $\mathbf{0.93}$ & $\mathbf{0.88}$ & $\mathbf{0.79}$\\
& Pedestrian & --- & $\mathbf{26.5}$ & $\mathbf{0.40}$ & $\mathbf{0.31}$ & $0.16$ \\

\hline
\end{tabular}
}
\end{center}
\vspace{-12pt}
\end{table}

\subsubsection{Perceptual quality evaluation}
\label{sec:percep_eval}
To quantify the perceptual quality of the results, we use the FID~\cite{heusel2017gans} between the composite images and the input. Moreover, we conduct a user study by performing A/B test to verify the realism of our approach. A pair of images composed upon the same input scene image, one by SAC-GAN and one by the baseline, are shown to 13 computer science graduate students. The users are told that the two images in a pair are composed by inserting an object into an existing scene image and are asked to choose which composition result is more realistic. Also, the users were asked to focus on the structure consistency between the inserted objects and the background instead of factors such as image resolution or color consistency. Table~\ref{tab:perceptual_iou} (left) reports the quantitative results for the car and pedestrian insertion. Both A/B test and FID values show our method clearly outperforms the four baselines. Note that the inferior performance for the Instance Insertion based results may be affected by the image synthesis algorithm used. A better semantic to real image translation method may increase the perceptual quality of final results. On the other hand, it also partially reflects the preference of users between the synthesis-based and direct object composition. In addition, we compare the computational cost of the direct composition methods in terms of inference time and the number of model parameters. It can be observed in Table \ref{tab:cost-para} that SAC-GAN only takes a reasonable longer inference time with a comparable network complexity.

We also show qualitative results in Fig.~\ref{fig:compare}, including visual composition images and semantic layouts. For ST-GAN and CompGAN, they may generate unstable locations and scales for the inserted objects and the objects tend to be distorted. For PlaceNet, it generates stable scales, yet produces undesirable locations because of the lack of semantic constraints. In comparison, our method consistently generates more plausible object transformation, since it learns the semantics and structure of the scene and object for composition. For example, the locations predicted by SAC-GAN appear to be more lane-aware than PlaceNet, even though there is no explicit annotation for lanes. The reason is that the shape of some masks (e.g., the road and sidewalk) implicitly contains the direction of the lane. By ensuring semantic coherence via adversarial layout loss, our results can respect the lanes to a certain degree. Note that the object scale of PlaceNet and SAC-GAN in Fig.~\ref{fig:compare} may look small. This is actually due to the car instance selected for self-supervised training needs to be intact, while such cars usually appear at relative far locations and their GT scales are also small for the Cityscapes dataset. For self-supervised methods like PlaceNet and ours, the object scale will vary depending on the datasets, e.g., the scales generally look fine for chair composition on 3D-FUTURE~\cite{fu20213d}. How to retain the benefits for self-training while being able to generate novel scales will be an interesting future work.

\setlength{\tabcolsep}{4pt}
\begin{table}[t]
\begin{center}
\caption{Comparison of computational cost and model parameters.}
\label{tab:cost-para}
\scalebox{0.92}{
\begin{tabular}{lcc}
\hline
Method & Inference time (sec) & Parameter number  \\
\hline
ST-GAN~\cite{lin2018st} & $0.42$ & 31.1M \\
\hline
CompGAN~\cite{azadi2020compositional} &  $1.21$ & 33.7M \\
\hline
PlaceNet~\cite{zhang2020learning} & $1.80$ & 37.6M \\
\hline
SAC-GAN & $1.91$ & 37.6M \\
\hline
\end{tabular}
}
\end{center}
\vspace{-12pt}
\end{table}

\begin{figure*}[t]
\begin{center}
{\centering\includegraphics[width=0.99\linewidth]{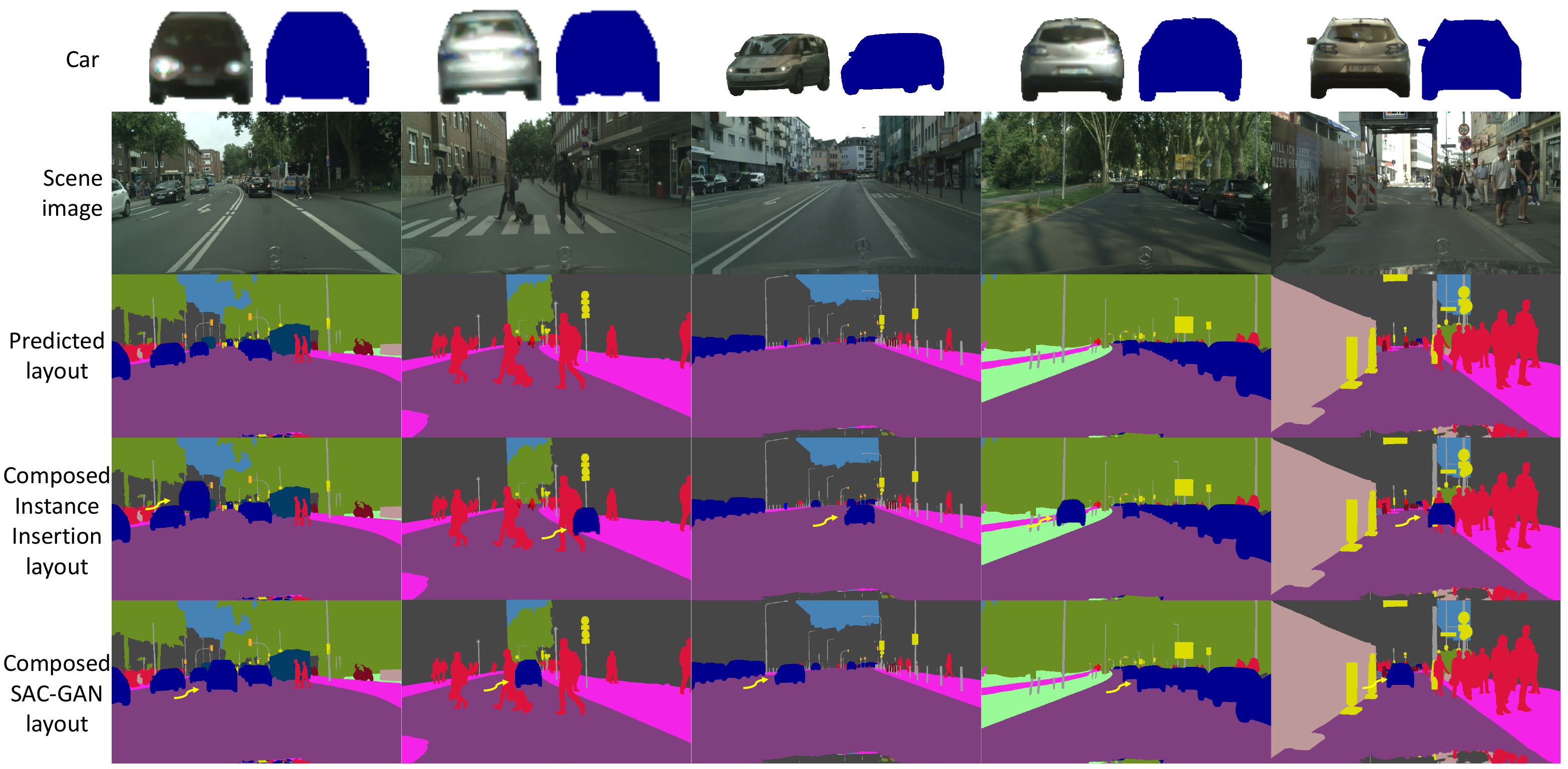}
}\\
\vspace{-15pt}
\caption{Qualitative comparison of SAC-GAN and Instance Insertion \cite{lee2018context} for inserting car masks into semantic layout images from Cityscapes.}
\label{fig:compare_semantic}
\end{center}
\end{figure*}

\begin{figure*}[h]
\begin{tabular}{ll}
\includegraphics[scale=0.60]{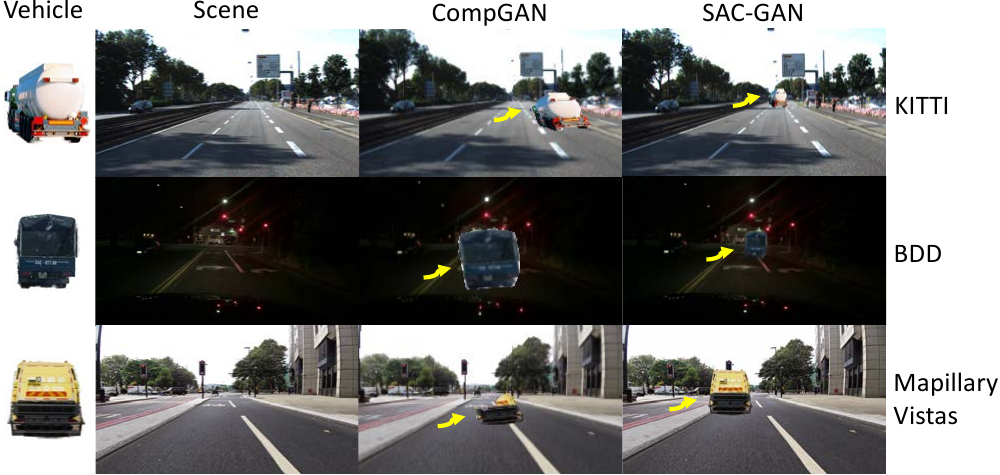}
&
\includegraphics[scale=0.5]{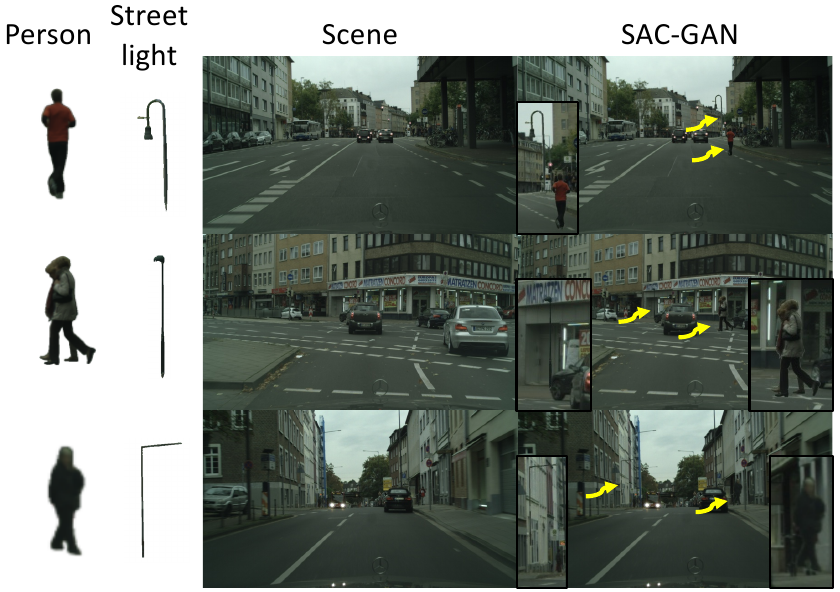}
\end{tabular}
\caption{Left: Cityscapes-trained networks (SAC-GAN vs.~CompGAN) applied to KITTI~\cite{geiger2013vision}, BDD~\cite{yu2020bdd100k}, and Mapillary Vistas~\cite{neuhold2017mapillary}, while the input object patches were from Google images. 
Right: Inserting a pedestrian and {\em then\/} a street light into the {\em same\/} background scene image by SAC-GAN.}
\label{fig:generalize_person}
\end{figure*}

\subsubsection{Comparison on semantic insertion}
\label{sec:semantic_eval}

To further evaluate how SAC-GAN performs on the semantic level, we perform quantitative and qualitative experiments of inserting the car masks into the semantic layout images of the Cityscapes using SAC-GAN and Instance Insertion. Table~\ref{tab:compare_semantic} shows the quantitative comparison on the FID values of the composite semantic layout images.
Evaluations are provided for both the insertion on GT and the semantic layout predicted by the off-the-shelf semantic segmentation model HMSA \cite{tao2020hierarchical}. It should be noted that the FID values between the GT-based and prediction-based composite layout are not directly comparable since the underlying distribution of GT and predicted layout images are different, while SAC-GAN can obtain composition with lower FID values when compared within each case. Fig.~\ref{fig:compare_semantic} shows the qualitative comparison for inserting the input car mask into the predicted semantic layout image. Both the quantitative and qualitative results show the superiority of SAC-GAN for composing objects at the semantic level.

\subsubsection{Comparison via object detection}
To validate the consistency between scene layouts and objects, we follow \cite{lee2018context} and use an object detector YOLOv5~\cite{glenn_jocher_2021_5563715} on the synthesized or composed images from Instance Insertion, ST-GAN, CompoGAN, PlaceNet and SAC-GAN. As YOLOv5 is context-aware, the detection degrades if the object is placed in an implausible region. We compute the precision at certain IoU threshold for evaluating the detection performance. From Table~\ref{tab:perceptual_iou} (right), the results show our method significantly outperforms the baselines in most cases. One exception is for pedestrian insertion with $\mathrm{IoU}_{85}$, the Instance Insertion performs slightly better, which may be due to they can generate pedestrians with larger sizes.

\begin{table}[t]
\begin{center}
\caption{Comparison on FID values of the composite semantic layout images obtained by inserting car masks on Cityscapes. The predicted layouts are generated from the pre-trained semantic segmentation model~\cite{tao2020hierarchical}. }
\label{tab:compare_semantic}
\scalebox{0.92}{
\begin{tabular}{lcc}
\hline
Method & FID$\downarrow$ (Predicted layout) & FID$\downarrow$ (GT layout)\\
\hline
Instance Insertion \cite{lee2018context} & $26.13$ & $30.23$\\
\hline
SAC-GAN & $\mathbf{20.71}$ & $\mathbf{24.57}$ \\
\hline
\end{tabular}
}
\end{center}
\vspace{-12pt}
\end{table}

\subsubsection{Generalizability}
We test the generalizability of our network from several aspects. First, we apply SAC-GAN trained on Cityscapes cars to compose Google searched vehicles on scene images from KITTI~\cite{geiger2013vision}, BDD~\cite{yu2020bdd100k}, and Mapillary Vistas~\cite{neuhold2017mapillary} datasets. Specifically, we have collected 9,071 truck, and 4,389 bus images via Google search and processed each image with object masks for qualitative testing. These images contain a large number of rare vehicles which are valuable for data augmentation and could be found through our project page. From Fig.~\ref{fig:generalize_person} (left), SAC-GAN can produce better results than CompGAN~\cite{azadi2020compositional} even the scene layouts and vehicles are different from training dataset. The superior performance of our network can be attributed to its awareness to scene semantics and structures.
In Fig.~\ref{fig:generalize_person} (right), we show our method is generally applicable to other object categories, such as pedestrians and street lights.

\setlength{\tabcolsep}{3pt}
\begin{table}
\begin{center}
\caption{IoUs of car, truck, and bus categories from two semantic segmentation models trained on original vs.~augmented (with Cityscapes cars) training set and evaluated on Cityscapes testing set.}
\label{tab:augmentation}
\scalebox{0.95}{
\begin{tabular}{lcccccc}
\hline
Data & \multicolumn{3}{c}{PSPNet~\cite{zhao2017pyramid}} & \multicolumn{3}{c}{DeepLabv3~\cite{chen2017rethinking}}  \\
& Car & Truck & Bus & Car & Truck & Bus \\
\hline
Original & $0.931$  & $\mathbf{0.476}$ &  $0.624$ & $0.874$ & $0.698$ &  $0.803$\\
\hline
Original + ST-GAN~\cite{lin2018st} & $0.928$ & $0.362$ & $0.624$  &  $0.877$ & $0.698$  & $\mathbf{0.810}$ \\
\hline
Original + CompGAN~\cite{azadi2020compositional} &  $0.938$ & $0.459$ & $0.755$   & $0.880$ & $\mathbf{0.704}$  & $0.798$ \\
\hline
Original + PlaceNet~\cite{zhang2020learning} & $0.934$ & $0.373$  & $\mathbf{0.773}$  &  $0.883$ & $0.664$  & $0.790$ \\
\hline
Original + SAC-GAN & $\mathbf{0.943}$ & $0.338$ & $0.746$  & $\mathbf{0.886}$ & $0.653$  & $0.792$\\
\hline
\end{tabular}
}
\end{center}
\vspace{-10pt}
\end{table}

\begin{figure*}[t]
\begin{center}
{\centering\includegraphics[width=0.99\linewidth]{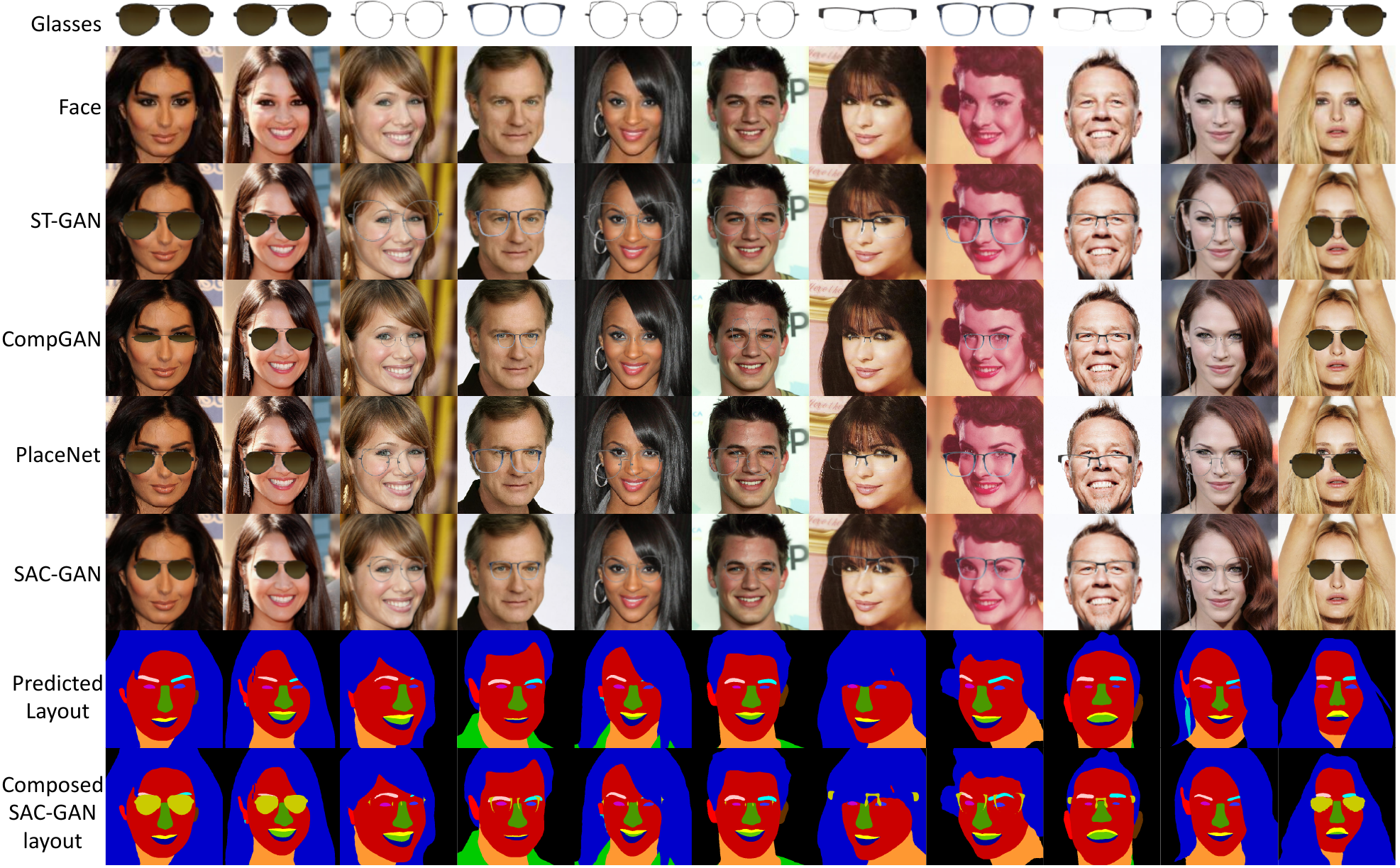}
}\\
\vspace{-12pt}
\caption{
Qualitative comparison of glasses composition with ST-GAN~\cite{lin2018st}, CompGAN~\cite{azadi2020compositional} and PlaceNet~\cite{zhang2020learning} on CelebAHQ dataset. Comparing with ST-GAN and CompGAN, SAC-GAN produces better object scales. As PlaceNet tends to have mode collapse with repetitive learned translation and scaling parameters, SAC-GAN is more stable with the additional VAE structure. 
We also show the semantic layout before and after our glasses composition. In general, SAC-GAN produces more satisfying location and scaling of the glasses than others. 
}
\label{fig:celeba}
\end{center}
\end{figure*}

\subsection{Application}
\label{sec:app}
In this section, we show the applications of SAC-GAN in road data augmentation, facial composition and indoor scene composition.
The main goal is to show the potential and the generalizability of SAC-GAN, and its superiority comparing to the learning-based composition baselines, but not aiming to propose the best solution for each task.

\subsubsection{Road data augmentation}
We investigate whether the composed images by SAC-GAN can augment current datasets to improve the performance of learning tasks in computer vision, such as semantic segmentation. Specifically, we augment the scene images from the training set of Cityscapes with one particular vehicle class and evaluate whether the semantic segmentation performance could be boosted for this class. We try to evaluate various options of image composition by using object patches from Cityscapes itself and from Google collected objects.

First, following the experimental setup in PlaceNet~\cite{zhang2020learning}, we augment each training scene with one of 2,975 Cityscapes-extracted cars. Then, we train semantic segmentation models PSPNet~\cite{zhao2017pyramid} and DeepLabv3~\cite{chen2017rethinking} respectively on data augmented with different methods and evaluate on the testing data using IoU metric. In Table~\ref{tab:augmentation}, the networks trained on the SAC-GAN augmented data show slight performance increase than others for the augmented car class, but underperform for the truck and bus classes. The reasons may be: 1) the segmentation performance for the car class is already high and it is difficult for the networks to be further improved; 2) augmentation on the car class which already contains more instances than truck or bus (e.g., over ten thousands vs. a few hundreds) will lead to further data imbalance and such augmentation can negatively impact the performance on classes with fewer observations.

\setlength{\tabcolsep}{3pt}
\begin{table}[t]
\begin{center}
\caption{IoUs of car, truck, and bus categories from two semantic segmentation models trained on original vs.~augmented (with Google trucks) training set and evaluated on Cityscapes testing set.}
\label{tab:augmentation2}
\scalebox{0.95}{
\begin{tabular}{lcccccc}
\hline
Data & \multicolumn{3}{c}{PSPNet~\cite{zhao2017pyramid}} & \multicolumn{3}{c}{DeepLabv3~\cite{chen2017rethinking}}  \\
& Car & Truck & Bus & Car & Truck & Bus \\
\hline
Original & $0.931$  & $0.476$ & $0.624$ & $0.874$ & $0.698$ &  $0.803$\\
\hline
Original + ST-GAN~\cite{lin2018st} & $0.953$ & $0.703$ & $\mathbf{0.800}$  &  $0.871$ & $0.597$  & $0.763$ \\
\hline
Original + CompGAN~\cite{azadi2020compositional} &  $0.918$ & $0.500$ & $0.431$   & $0.865$ & $0.711$  & $\mathbf{0.827}$ \\
\hline
Original + PlaceNet~\cite{zhang2020learning} & $0.910$ & $0.431$  & $0.445$  &  $0.855$ & $0.412$  & $0.589$ \\
\hline
Original + SAC-GAN & $\mathbf{0.954}$ & $\mathbf{0.732}$ & $0.752$  & $\mathbf{0.874}$ & $\mathbf{0.782}$  & $0.801$\\
\hline
\end{tabular}
}
\end{center}
\vspace{-10pt}
\end{table}

\setlength{\tabcolsep}{3pt}
\begin{table}[t]
\begin{center}
\caption{IoUs of car, truck, and bus categories from two semantic segmentation models trained on original vs. augmented (with Google buses) training set and evaluated on Cityscapes testing set.}
\label{tab:augmentation3}
\scalebox{0.95}{
\begin{tabular}{lcccccc}
\hline
Data & \multicolumn{3}{c}{PSPNet~\cite{zhao2017pyramid}} & \multicolumn{3}{c}{DeepLabv3~\cite{chen2017rethinking}}  \\
& Car & Truck & Bus & Car & Truck & Bus \\
\hline
Original & $0.931$  & $0.476$ &  $0.624$ & $0.874$ & $0.698$ &  $0.803$\\
\hline
Original + ST-GAN~\cite{lin2018st} & $0.924$ & $0.523$ & $0.694$  &  $\mathbf{0.927}$ & $0.542$  & $0.670$ \\
\hline
Original + CompGAN~\cite{azadi2020compositional} &  $0.949$ & $0.719$ & $0.757$   & $0.871$ & $0.620$  & $0.681$ \\
\hline
Original + PlaceNet~\cite{zhang2020learning} & $0.919$ & $0.491$  & $0.574$  &  $0.867$ & $0.639$  & $0.804$ \\
\hline
Original + SAC-GAN & $\mathbf{0.952}$ & $\mathbf{0.743}$ & $\mathbf{0.798}$  & $0.872$ & $\mathbf{0.772}$  & $\mathbf{0.824}$\\
\hline
\end{tabular}
}
\end{center}
\vspace{-10pt}
\end{table}

\begin{figure*}[t]
\begin{center}
{\centering\includegraphics[width=0.95\linewidth]{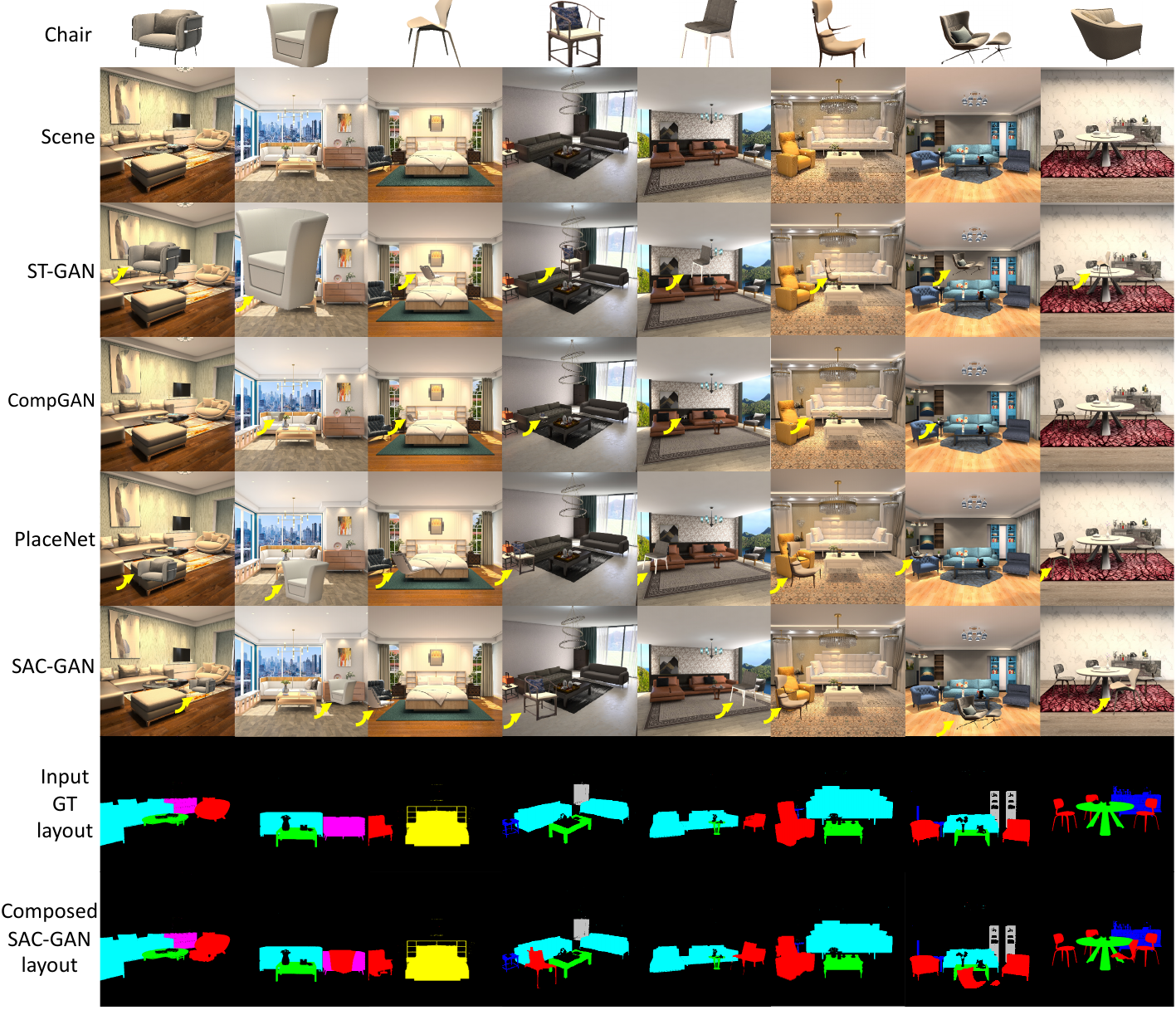}
}\\
\vspace{-10pt}
\caption{
Indoor scene chair composition results. Compared to ST-GAN, CompGAN and PlaceNet, SAC-GAN generates more realistic composites. Note, the iterative warp update in ST-GAN may be affected by the initial inaccurate prediction, and it is hard to generalize CompGAN to diverse indoor scenes. }
\label{fig:chair}
\end{center}
\end{figure*}

Next, we perform data augmentation for the truck and bus classes separately, using SAC-GAN and other methods. Unlike the car augmentation experiment, we use trucks and buses from our Google collected images to test whether the external data can be used for augmenting the specified class. In detail, we randomly select 100 Google collected and processed truck/bus object patches and compose one truck/bus into each image of the Cityscapes training set. Segmentation models are trained in the same way as above.
As shown in Table~\ref{tab:augmentation2} and Table~\ref{tab:augmentation3}, even based on a small set of Google truck/bus patches, the semantic segmentation performance for the augmented class can be significantly improved with the training data augmented by SAC-GAN, while the data augmented by other methods may degrade the network's performance. In these experiments, it is also interesting to see the performance of unaugmented classes (e.g., the car and bus in Table~\ref{tab:augmentation2}) may also increase. This may be due to the network being able to boost its overall ability if a confusing or difficult class (i.e., the truck in Table~\ref{tab:augmentation2}) can be better segmented.

From the above experiments, it can be seen SAC-GAN is more effective than other compared baselines in data augmentation via image composition. The reason is the plausibility of the location and the scale of the composed object patches will affect the learning of semantic segmentation models. On the other hand, it is worth mentioning that we only show the potential of SAC-GAN for its application in data augmentation and demonstrate its superiority comparing to other image composition methods for this application. How to design a better composition-based data augmentation scheme, e.g., whether external data should be used and what are the more proper percentages for each class to be augmented, is beyond the scope of this paper and it will be a promising direction to investigate in the future.

\subsubsection{Facial composition}
We show the potential of SAC-GAN on composing glasses on human faces. We utilize CelebAHQ~\cite{liu2015faceattributes} dataset and separate the data into two sets: people with and without glasses. We train SAC-GAN in the same self-supervised manner by performing glass cropping on the people with glasses set. Fig.~\ref{fig:celeba} shows the qualitative comparison of SAC-GAN with ST-GAN, CompGAN and PlaceNet, as well as the generated semantic layouts. Comparing with ST-GAN and CompGAN, SAC-GAN generates better object scales. As PlaceNet tends to have mode collapse with repetitive learned translation and scaling parameters, SAC-GAN is more stable with the VAE-GAN structure. However, it can also be observed that some glasses may not match well with the faces (the 7th and 8th column in Fig.~\ref{fig:celeba}). The main reason is when performing self-training, the glasses and faces in the training images have the same orientations, but when testing, the input glasses are from the canonical view and some faces are not looking front. Therefore, the result may degrade since our image composition only considers translation and scaling for transformation learning based on self-supervised paired data. How to enable more flexible and more plausible image composition that allows affine transformation will be an interesting future work. In Table~\ref{tab:quantitative_glasses_chair} left, we quantitatively evaluate the composition quality by human A/B test and FID, and SAC-GAN outperforms the other three methods for both metrics by a wide margin.  

\floatsetup[table]{capposition=top}
\begin{table}[t]
\begin{center}
\caption{Quantitative comparisons of perceptual quality evaluation results for wearing glasses on faces and inserting chairs on indoor scenes. }
\label{tab:quantitative_glasses_chair}
\begin{tabular}{lccc|ccc}
\hline
Method & Type & HS ($\%$)$\uparrow$ & FID$\downarrow$ & Type & HS ($\%$)$\uparrow$ & FID$\downarrow$ \\
\hline
ST-GAN~\cite{lin2018st} & glasses & $65.6$ & $72.4$ & chair & $97.5$ & $110.1$ \\
\hline
CompGAN~\cite{azadi2020compositional} & glasses & $66.3$ & $84.6$ & chair & $72.6$ & $33.9$ \\
\hline
PlaceNet~\cite{zhang2020learning} & glasses & $98.8$ & $87.0$ & chair & $45.1$ & $33.5$ \\
\hline
SAC-GAN & glasses & --- &  $\mathbf{66.9}$ & chair & --- & $\mathbf{32.2}$ \\
\hline
\end{tabular}
\end{center}
\vspace{-10pt}
\end{table}

\begin{figure*}[t]
\centering
\includegraphics[width=0.99\linewidth]{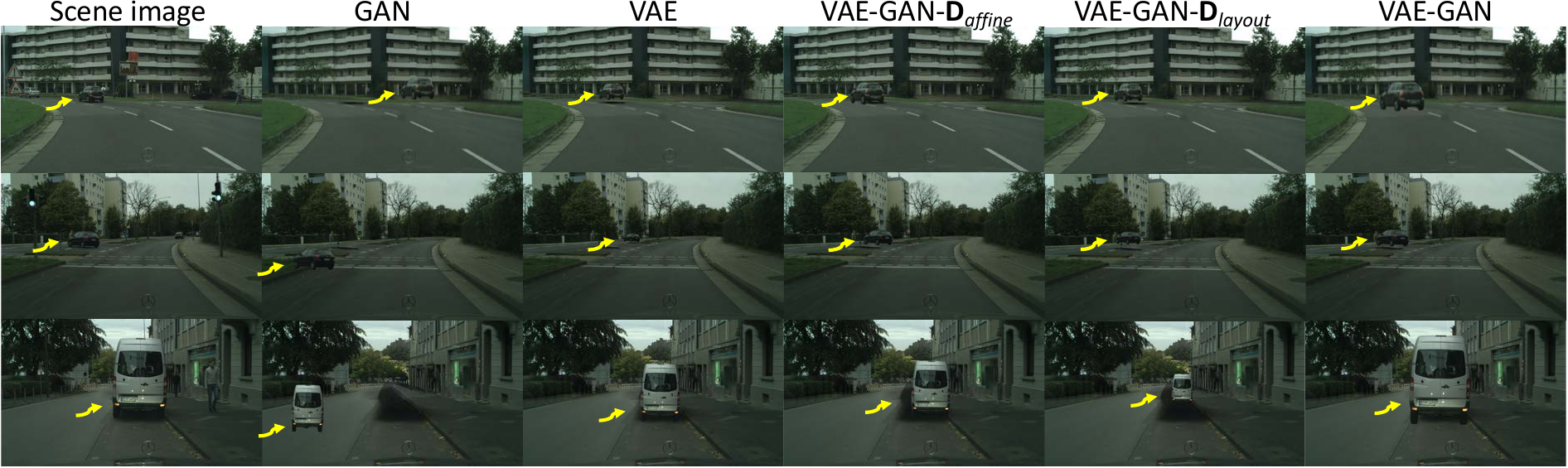}
\caption{Ablation study on loss functions. We compare the insertion location and size obtained from different variants of our method with the original object in the scene (first column). Note that the background scene is obtained by taking out the foreground object and performing inpainting to fill the holes. Please pay attention to the placed object instead of the inpainted region.}
\label{fig:vaegan}
\end{figure*}

\begin{figure}[t]
\begin{center}
{\centering\includegraphics[width=0.95\linewidth]{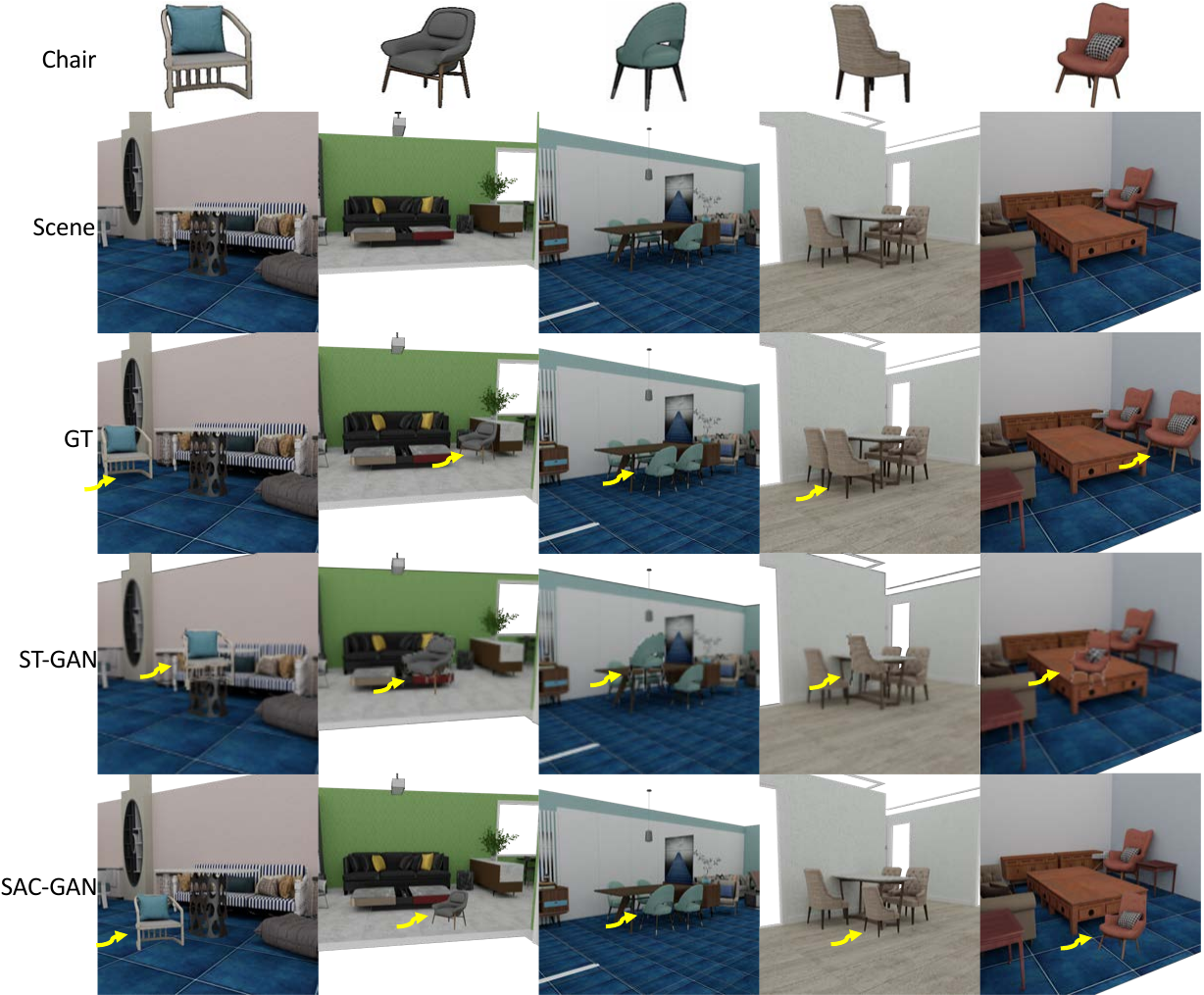}
}\\
\vspace{-10pt}
\caption{
Qualitative comparison for chair composition on 3D-FRONT~\cite{fu_iccv21} rendered images. In general, SAC-GAN produces more satisfying location and scaling of the chairs than ST-GAN. }
\label{fig:render}
\end{center}
\end{figure}

\subsubsection{Indoor scene composition}
We perform additional experiments on composing indoor-scene chair composition. For indoor scenes, we use 3D-FUTURE~\cite{fu20213d} dataset for training and testing. The dataset provides rendered images and the corresponding semantic labels, and we generate the semantic layouts by re-mapping the labels into seven major categories (e.g., chair, sofa etc) to focus on the overall scene semantics and structure. We train the model for compositing the chair category using the same training strategy as before, with
Fig.~\ref{fig:chair} comparing SAC-GAN and ST-GAN, CompGAN and PlaceNet results. In Table~\ref{tab:quantitative_glasses_chair} right, we quantitatively evaluate the composition quality, and SAC-GAN outperforms the state-of-the-art methods. We find that the iterative warp update scheme of ST-GAN may be affected by the initial prediction and cause error propagation if the first few steps are not updated correctly, while SAC-GAN is almost free from such failure cases. Similarly, CompGAN generates unstable scales and locations for more complex scenes like indoor scenes. For PlaceNet, since it dose not explicitly focus on the structural compatibility for the composite image as SAC-GAN does, the predicted location may be inplausible for the input chair with various orientations.

In addition to comparing object composition on 3D-FUTURE \cite{fu20213d} images, we also provide qualitative comparisons for chair composition on 3D-FRONT~\cite{fu_iccv21} rendered images to show the generalizability of SAC-GAN and ST-GAN. To provide a reference for a proper insertion, we render a pair of images for each scene: one GT image with all objects of the original scene and the other background image with a specified chair removed from the scene. Then, we perform chair composition by inserting the input chair into the background images using ST-GAN and SAC-GAN. The results in Fig. \ref{fig:render} show that ST-GAN tends to place the chair in the middle of the scene and ignore other surrounding objects, while SAC-GAN can generally insert the chair at plausible location.

\subsection{Ablation Study}
In the following, we perform ablation studies on car insertion to verify the effectiveness of different losses and designs in SAC-GAN. To ensure the evaluation is less affected by the data distribution of the training scenes, we re-train the SAC-GAN and compare the results on the more city-focused Cityscapes-Aachen subset.

\subsubsection{Effects of VAE-GAN}
In Fig.~\ref{fig:vaegan} and Table~\ref{tab:ablationa}, we compare our method with several variations: 
a) GAN loss $\mathcal{L}_{GAN}$ only;
b) VAE loss $\mathcal{L}_{VAE}$ only;
c) VAE and $\mathbf{D}_{\it{affine}}$ loss;
d) VAE and $\mathbf{D}_{layout}$ loss;
e) Full VAE-GAN loss $\mathcal{L}_{VAE-GAN}$.
Theoretically, for a), while the adversarial losses can help the network learn a diverse transformation, the result may be unstable. In addition, when mode collapse happens, the network will always output the same transformation. For b), although the network tries to produce proper transformations based on the latent distribution, it may fail to generate diverse object locations. Comparing to b), adding the $\mathbf{D}_{\it{affine}}$ loss in c) can bring more variations to the object locations and scales. Similarly, adding the $\mathbf{D}_{layout}$ loss in d) can ensure the network to respect the scene layouts. Lastly, with the full VAE-GAN loss, the results should be more diverse and plausible object transformations. The results in Fig.~\ref{fig:vaegan} confirm above analysis. The quantitative comparison in Table~\ref{tab:ablationa} also verifies the effectiveness of the proposed losses. Hence, our VAE-GAN architecture is essential to learn more realistic image-to-image composition.

\begin{table}[t]
\caption{Quantitative results of ablation study on effects of loss functions (the first four data columns), scene semantics (compared with RGB-based layout loss) and the object structure (last two columns).}
\setlength{\tabcolsep}{2.5pt}
\scalebox{0.82}{
\begin{tabular}{lccccccc}
\hline
Method & GAN & VAE & VAE-GAN & VAE-GAN & RGB-based & VAE-GAN  & VAE-GAN  \\
 & & & $D_{\it{affine}}$ only & $D_{layout}$ only & layout loss & w/o edge & Ours \\
\hline
FID$\downarrow$ & $108$ & $93.2$ & $96.9$ & $98.3$  & $96.2$ & $89.5$ & $\mathbf{88.4}$ \\
\hline
\end{tabular}
}
\label{tab:ablationa}
\end{table}

\subsubsection{Effects of semantics and structures}
\begin{figure}[]
\begin{center}
{\centering\includegraphics[width=0.98\linewidth]{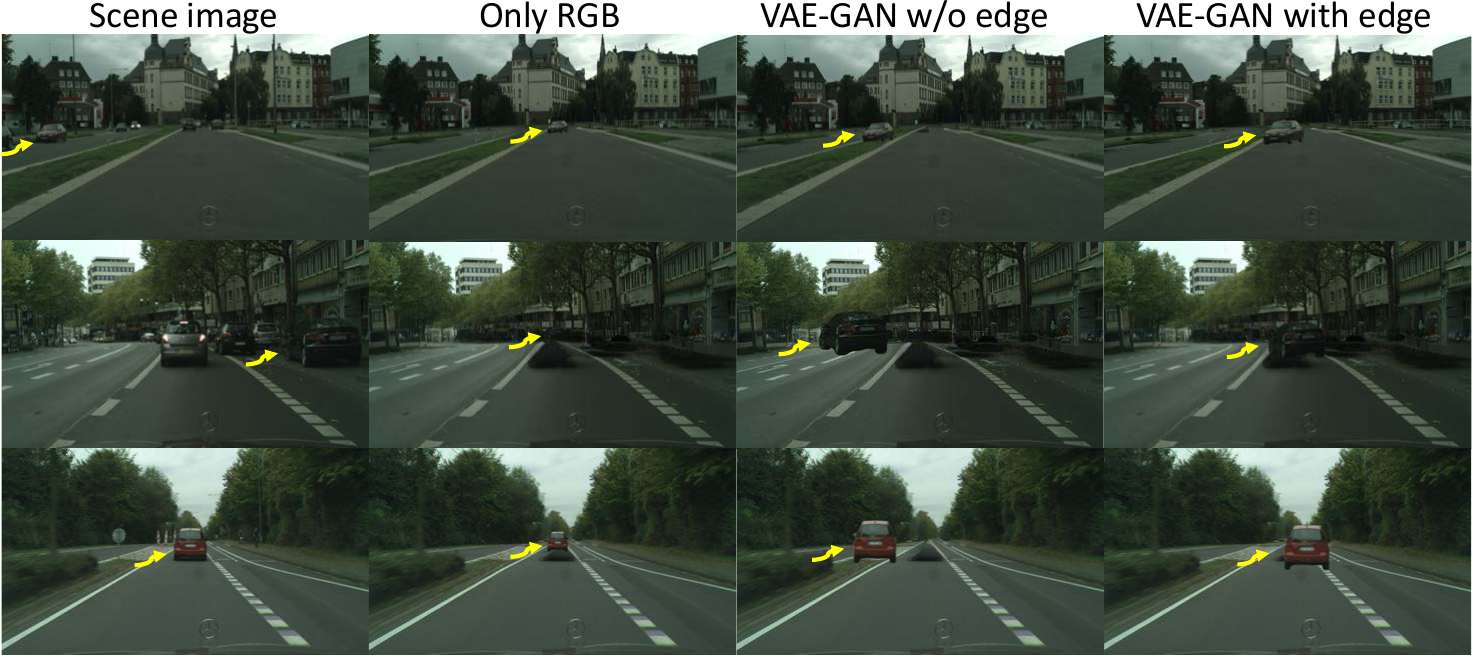}
}\\
\caption{
Ablation study on semantics and structures. Similar inpainting operation as in the study of losses is performed. The results of using RGB instead of the masks to compute layout loss (second column) and dropping the object edges in object encoder (third column) show inferior performance than our proposed design.}
\label{fig:edge}
\end{center}
\end{figure}

The effects of semantic layouts and object structures (i.e., edge maps) in training are shown in Fig.~\ref{fig:edge} and Table \ref{tab:ablationa} (last two columns). We compare our method with two variants: a) replacing the mask-based layout loss (Eq. \ref{eqn:07}) with pixel-wise loss computed from the paired composed RGB scene image and inpainted image; b) dropping the edge map from the object encoder. For a), the inpainted images are generated by Co-Mod~\cite{zhao2020large} by removing the target object in the scene. The model trained on the paired inpainted images and composed images shows similar performance with SAC-GAN, but the variant occasionally failed to learn the correct object locations and scales since the network becomes less-focused to scene semantics. For the variant b), we observe that the inserted objects are more likely to overlap regions outside the road or of improper sizes, which indicates the importance of the object edge maps when learning the object-scene relationships.

\section{Conclusion, Limitations, and Future Work}
\label{sec:conclusion}

\begin{figure}[t]
\centering
\includegraphics[width=0.98\linewidth]{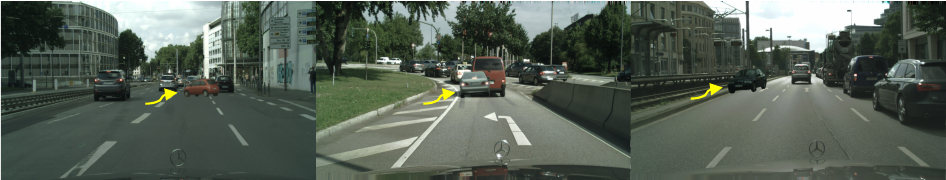}
\caption{Failure cases. From left to right: misaligned, occluded, and semantically incongruent vehicle insertions.}
\label{fig:failure}
\end{figure}

We present SAC-GAN, a novel self-supervised and adversarial network for an object-oriented image composition approach that can be used for semantic-aware image manipulation and augmentation. Compared to scene manipulation relying on 3D models, our purely image-space composition avoids the challenge of 3D reconstruction and can benefit from access to a much larger volume of data. On the flip side, altering object poses from a single patch is not possible without resorting to novel view synthesis \cite{chen2021geosim}, leading to unrealistic composite images with misaligned objects (Fig. \ref{fig:failure} left). Also, occlusions are not specifically avoided (Fig. \ref{fig:failure} middle), and our current method is not yet able to account for semantic constraints such as driving directions (Fig. \ref{fig:failure} right) and stylistic compatibility (see various results in Fig.~\ref{fig:chair}). Besides, although the locations of our car compositions can respect the lanes to a certain degree, it is difficult to predict the most appropriate locations among different lanes when the lane constraints are not explicitly learned.
Extending SAC-GAN with lane constraints obtained from lane detection is worth exploring. Currently, our method focuses on structural coherence, not pixel-level realism such as lighting or color harmonization. Although some post-refinement techniques such as image harmonization \cite{cong2020dovenet} and image-based shadow generation \cite{liu2020arshadowgan} may be used to improve the photorealism of composite images, directly integrating pixel-level quality control into SAC-GAN is of future interest.

Furthermore, there remains much to do in modelling the 2D transforms via VAE-GAN. Although as a GAN-based method, SAC-GAN is expected to have the ability to produce diversified results. However, our current object composition tends to only vary slightly on the predicted location and scale. This is due to the fact that VAE-GAN network is also a conditional generative model and is highly constrained by the input scene and object. With the current model trained on Cityscapes which only contains a not-so-large-scale number of usable images for self-supervised composition learning, it is possible the network mainly attempts to predict the most plausible transformation based on the limited object-scene relationships learned. Enriching the training data by introducing more images from other datasets such as BDD and Mapillary Vistas may enhance the diversity and plausibility of our results. In addition, modeling the semantic relations with a graph representation and incorporating the 3D inference may reduce current strong constraints on the 2D layout masks. Nevertheless, the VAE-GAN may still not generate plausible transforms when an object and a scene are not sufficiently composable. This poses an interesting question of how to learn a {\em composability score\/} to allow quick selection of composable image pairs rather than predicting arbitrary transformations for the incompatible input. Last but not least, to allow more flexible control over the image composition, learning the deep interaction between natural language and the image objects with recent vision-language models \cite{radford2021learning,wang2022cris, ding2022vlt} and then using the text prompts to guide the object insertion are promising future directions to explore.

\bibliographystyle{abbrv-doi}

\bibliography{bib}

\par\noindent 
\parbox[t]{\linewidth}{
\noindent\parpic{\includegraphics[height=1.5in,width=1in,clip,keepaspectratio]{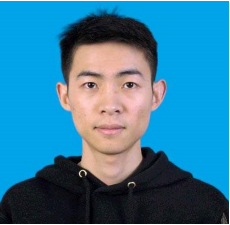}}
\noindent {\bf Hang Zhou}\
received his B.S. degree in 2015 from Shanghai University and Ph.D. in 2020 from the University of Science and Technology of China. Currently, he is a postdoctoral researcher at Simon Fraser University. His research interests include computer graphics, multimedia security and deep learning.}
\vspace{2\baselineskip}

\par\noindent 
\parbox[t]{\linewidth}{
\noindent\parpic{\includegraphics[height=1.5in,width=1in,clip,keepaspectratio]{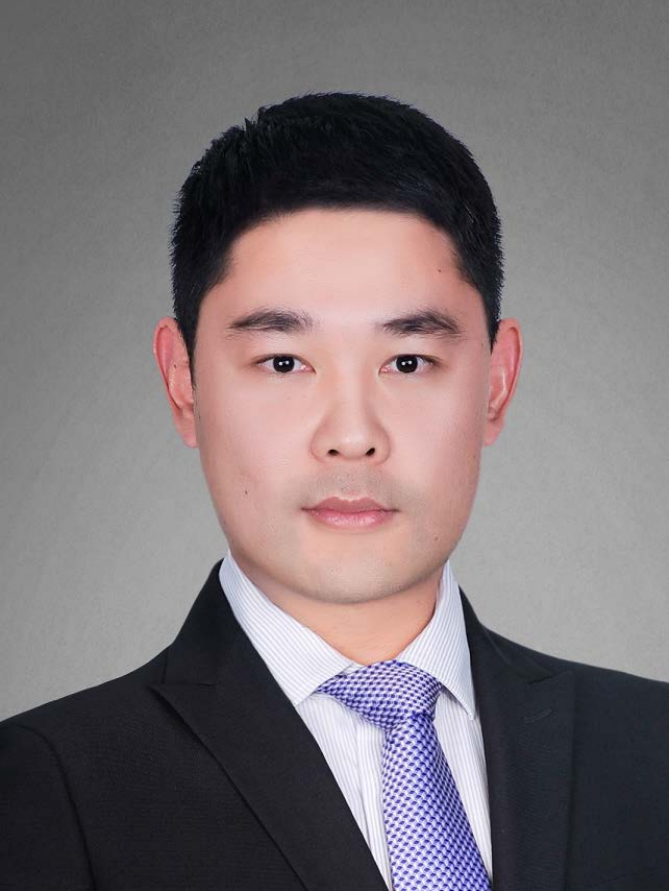}}
\noindent {\bf Rui Ma}\
is an associate professor in School of Artificial Intelligence, Jilin University, China. He obtained his PhD in 2018 from School of Computing Science, Simon Fraser University, Canada and his MSc and BSc from School of Mathematics, Jilin University, China. His research is in computer graphics, computer vision and artificial intelligence, with special interests in intelligent analysis, creation and application of visual content.}
\vspace{2\baselineskip}

\par\noindent 
\parbox[t]{\linewidth}{
\noindent\parpic{\includegraphics[height=1.5in,width=1in,clip,keepaspectratio]{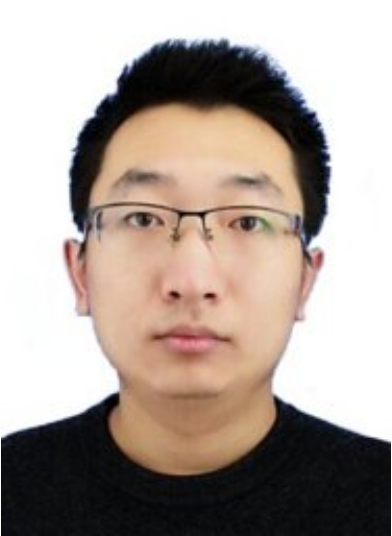}}
\noindent {\bf Ling-Xiao Zhang}\
received his Master of Engineering's degree in Computer Technology from Chinese Academy of Sciences in 2020. He is currently an assistant engineer in the Institute of Computing Technology, Chinese Academy of Sciences. His research interests include computer graphics and geometric processing.}
\vspace{2\baselineskip}

\par\noindent 
\parbox[t]{\linewidth}{
\noindent\parpic{\includegraphics[height=1.5in,width=1in,clip,keepaspectratio]{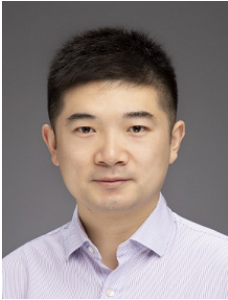}}
\noindent {\bf Lin Gao}\
received the BS degree in mathematics from Sichuan University and the PhD degree in computer science from Tsinghua University. He is currently an associate professor in Institute of Computing Technology, Chinese Academy of Sciences. His research interests include computer graphics, geometric modeling and processing.}
\vspace{2\baselineskip}

\par\noindent 
\parbox[t]{\linewidth}{
\noindent\parpic{\includegraphics[height=1.5in,width=1in,clip,keepaspectratio]{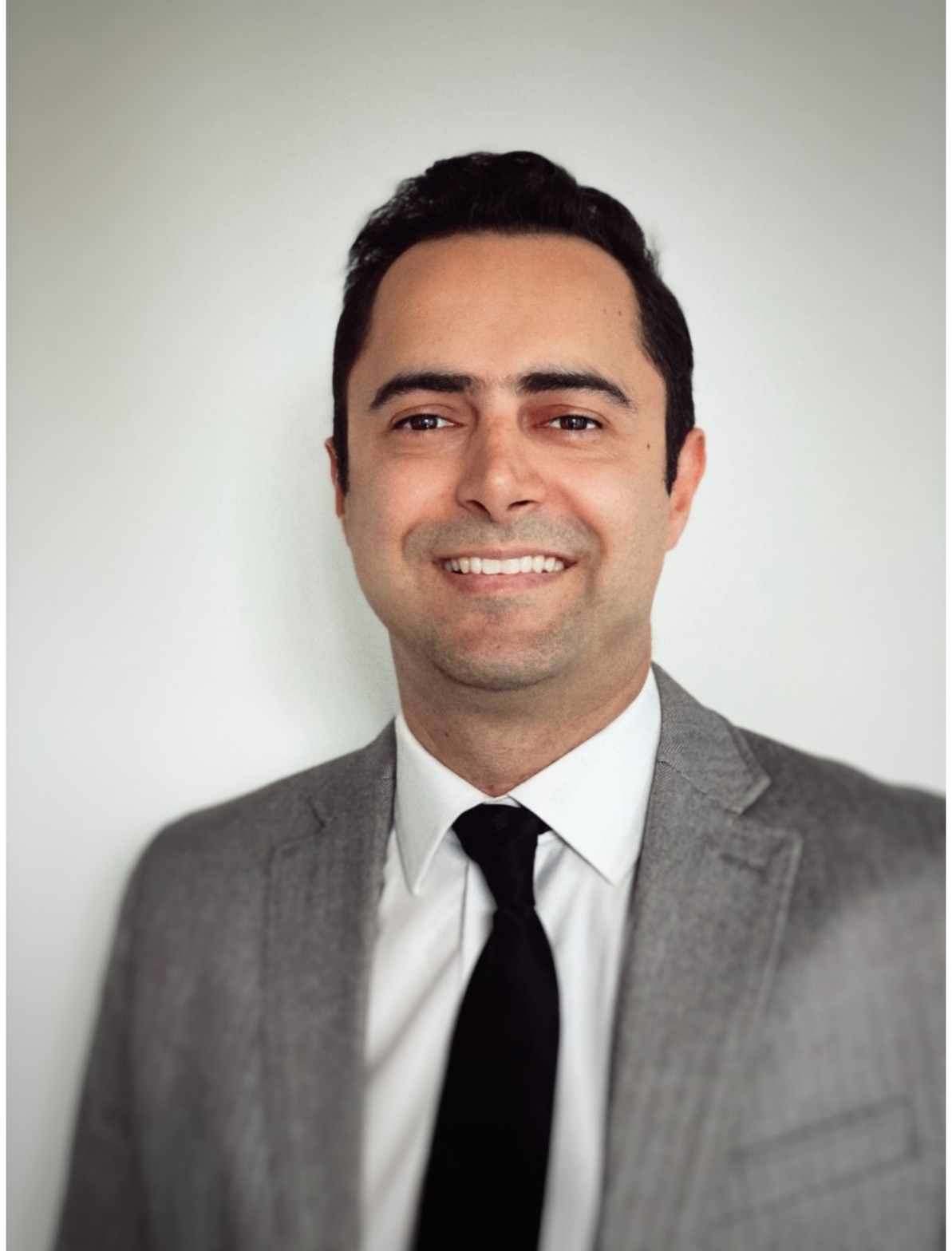}}
\noindent {\bf Ali Mahdavi-Amiri}\
is currently an assistant professor in GrUVi lab at Simon Fraser University, Canada. He received his PhD from the University of Calgary in 2015. His research area is geometric modeling and computational creativity, design, and fabrication. He has received several outstanding grants and awards including NSERC Postdoctoral Fellowship, J.B. Hyne Research Innovation Award, and Michael A. J. Sweeney Award for the best paper at Graphics Interface, 2013.}
\vspace{2\baselineskip}

\par\noindent 
\parbox[t]{\linewidth}{
\noindent\parpic{\includegraphics[height=1.5in,width=1in,clip,keepaspectratio]{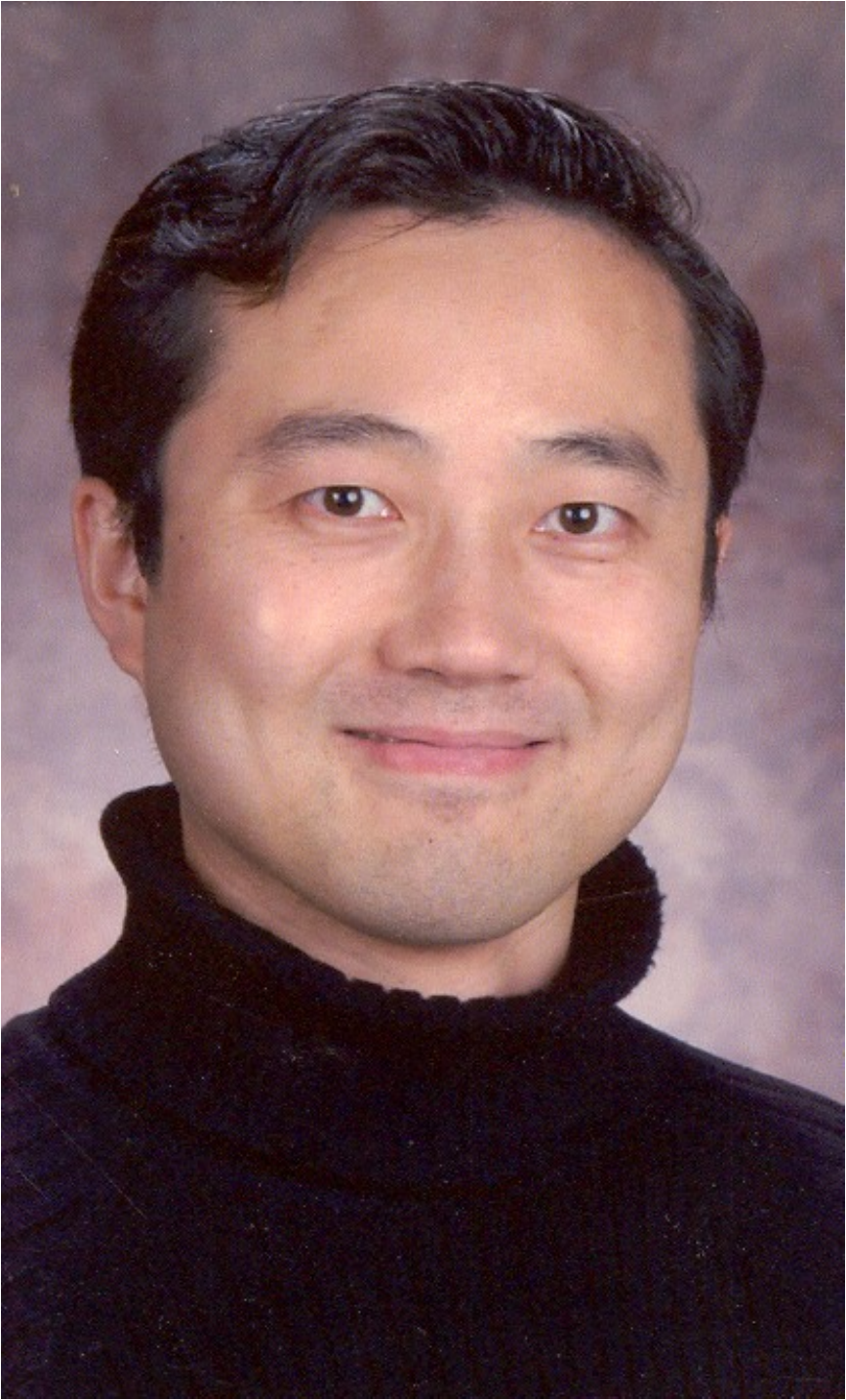}}
\noindent {\bf Hao Zhang}\
is a full professor in the School of Computing Science at Simon Fraser University (SFU), Canada, where he directs the graphics (GrUVi) lab. He obtained his Ph.D. from the Dynamic Graphics Project (DGP), University of Toronto, and M.Math. and B.Math degrees from the University of Waterloo, all in computer science. Richard's research is in computer graphics with a focus on geometric deep learning, 3D shape analysis and content creation, 3D vision, and computational design and fabrication.}

\end{document}